\documentclass[3p]{elsarticle}
\usepackage{framed,multirow}
\usepackage{amsthm}
\usepackage{latexsym}
\usepackage{amssymb,amsmath,mathtools,amsfonts}
\usepackage{graphicx,subfig,version,color}
\usepackage{xcolor}
\usepackage{subcaption}
\usepackage{caption}
\usepackage{cases}
\usepackage{exscale}
\usepackage{relsize}
\usepackage{hyperref}
\hypersetup{
colorlinks=true,
linkcolor=black
}
\numberwithin{figure}{section}
\numberwithin{equation}{section}
\newtheorem{thm}{Theorem}[section]

\usepackage{url}
\usepackage{xcolor}
\definecolor{newcolor}{rgb}{.8,.349,.1}
\biboptions{sort&compress}
\graphicspath{{figs/}}
\usepackage{bm}
\usepackage{booktabs}
\usepackage{longtable}
\usepackage{caption}
\usepackage[ruled,vlined,linesnumbered]{algorithm2e}
\usepackage{algpseudocode}%

\makeatletter
\newcommand{\removelatexerror}{\let\@latex@error\@gobble}
\makeatother
\usepackage{listings}%
\usepackage{makecell}
\definecolor{newcolor}{rgb}{.8,.349,.1}

\DeclareMathOperator{\diag}{diag}
\allowdisplaybreaks[4]
\graphicspath{{figs/}}
\catcode`\@=11
\@addtoreset{equation}{section} % Makes \section reset 'equation' counter.
\usepackage{url}
\usepackage{xcolor}
\definecolor{newcolor}{rgb}{.8,.349,.1}
\biboptions{sort&compress}
\graphicspath{{figs/}}
\newcommand{\Bone}{{\bm 1}}
\newcommand{\e}{{\mathrm{e}}}
\newcommand{\p}{{\mathrm{p}}}
\newcommand{\h}{{\mathrm{h}}}
\newcommand{\bx}{\boldsymbol{x}}
\newcommand{\tF}{{\rm F}}

\newcommand{\pt}{\partial_{t}}
\newcommand{\ptt}{\partial_{tt}}
\newcommand{\pxi}{\partial_{x_i}}
\newcommand{\pxii}{\partial_{x_ix_i}}
\newcommand{\R}{\mathcal{R}}
\newcommand{\T}{\mathcal{T}}
\newcommand{\Q}{\mathcal{Q}}
\newcommand{\Je}{\mathcal{J}^\mathrm{e}}
\newcommand{\Jp}{\mathcal{J}^\mathrm{p}}
\newcommand{\Jh}{\mathcal{J}^\mathrm{h}}
\newcommand{\Jse}{{\mathcal{J}^\mathrm{e}_{\rm{S}}}}
\newcommand{\Jsp}{{\mathcal{J}^\mathrm{p}_{\rm{S}}}}
\newcommand{\Jsh}{{\mathcal{J}^\mathrm{h}_{\rm{S}}}}
\newcommand{\no}{\nonumber}

\newcommand{\hEltwo}{\hat{\mathcal{E}}_{\ell^2}}
\newcommand{\Bt}{\Bone^\top}

\begin{document}
\sloppy
\begin{frontmatter}
\title{Layer Separation Deep Learning Model with Auxiliary Variables for Partial Differential Equations}
\author[1]{Yaru Liu}
\ead{yaruliu@std.uestc.edu.cn}
\author[1]{Yiqi Gu\corref{cor1}}
\ead{yiqigu@uestc.edu.cn}
\cortext[cor1]{Corresponding author}
\address[1]{School of Mathematical Sciences, University of Electronic Science and Technology of China, Sichuan, China}
\begin{abstract}

In this paper, we propose a new optimization framework, the layer separation (LySep) model, to improve the deep learning-based methods in solving partial differential equations. Due to the highly non-convex nature of the loss function in deep learning, existing optimization algorithms often converge to suboptimal local minima or suffer from gradient explosion or vanishing, resulting in poor performance. To address these issues, we introduce auxiliary variables to separate the layers of deep neural networks. Specifically, the output and its derivatives of each layer are represented by auxiliary variables, effectively decomposing the deep architecture into a series of shallow architectures. New loss functions with auxiliary variables are established, in which only variables from two neighboring layers are coupled. Corresponding algorithms based on alternating directions are developed, where many variables can be updated optimally in closed forms. Moreover, we provide theoretical analyses demonstrating the consistency between the LySep model and the original deep model. High-dimensional numerical results validate our theory and demonstrate the advantages of LySep in minimizing loss and reducing solution error.
\end{abstract}
%\begin{subclass}
%\subclass{68T07,  90C26, 65M99, 65N99}
%\end{subclass}
\end{frontmatter}
	
\section{Introduction}
Partial differential equations (PDEs) are widely used in physics, engineering, finance, and other fields \cite{Yserentant2005,Lee2002,Ehrhardt2008}. They can describe and explain changes in complex systems, such as the motion of objects, changes in electromagnetic fields, the evolution of biological populations, and other phenomena. In the past decades, numerical methods for the solution of PDEs have been developed significantly. Traditional numerical methods (e.g., finite difference method, finite element method, and spectral method) usually transform differential equations into algebraic systems through discretization. Although these methods have a strict theoretical foundation and high computational accuracy, they are limited by the curse of dimensionality, which results in an exponential increase in storage requirements as the dimensionality increases. Thus, it is difficult to solve high-dimensional problems effectively by these methods.

In recent years, a series of methods based on deep neural networks (DNNs) (e.g., physics-informed neural networks (PINNs)\cite{Cai2022,Cai2021,Jagtap2020,Mao2020,Pang2019,Raissi2019,Rao2020}, deep Ritz methods \cite{Yu2018,Gu2021-2}, and weak adversarial networks (WANs) \cite{Zang2020,Bao2024}, etc.) have been proposed for solving high-dimensional PDEs. Basically, the unknown solutions of PDEs are approximated by a DNN, which is trained by minimizing a specific loss function. Due to the special nonlinear structure of DNNs, the number of free parameters grows moderately with dimension, making it possible to realize high-dimensional approximations with limited storage. As an example, we consider the following PDE with Dirichlet boundary condition,
 \begin{equation}\label{01}
 \begin{aligned}
     \mathcal{D}[u,\bx]&=f(\bx),\quad \bx\in\Omega,\\
     u(\bx)&=g(\bx),\quad \bx\in\partial\Omega,
 \end{aligned}
 \end{equation}
where $\mathcal{D}$ is some differential operator; $u$ is the unknown solution; $f$ and $g$ are given function; $\Omega$ is a $d$-dimensional domain and $\partial\Omega$ is its boundary. In the PINN model, a neural network $\phi(\bx;\theta)$ with free parameters $\theta$ is taken to approximate the solution $u(\bx)$, which is trained through the following optimization problem
\begin{equation}\label{02}
\min_{\theta}\frac{1}{N_1}\sum_{n=1}^{N_1}|\mathcal{D}[\phi(\bx_n^{(1)};\theta),\bx_n^{(1)}]-f(\bx_n^{(1)})|^2+\frac{\mu}{N_2}\sum_{n=1}^{N_2}|\phi(\bx_n^{(2)};\theta)-g(\bx_n^{(2)})|^2,
\end{equation}
where $\mu\geq 0$ is a scalar weight; $\{x_n^{(1)}\}_{n=1}^{N_1}$ and $\{x_n^{(2)}\}_{n=1}^{N_2}$ are the datasets of feature points in $\Omega$ and on $\partial\Omega$, respectively. Note that the first term is the PDE residual and the second is the boundary condition error. One aims to find the optimal neural network $\phi(\bx;\theta)$ as an approximate solution of \eqref{01} by solving \eqref{02}. 
 
Due to the highly non-convex nature of the loss function, commonly used optimizers (e.g., stochastic gradient descent) may converge to bad local minima if the initialization or hyperparameters are not properly assigned. Although it is theoretically demonstrated that the global optima of \eqref{02} can be attained in the over-parameterized regime \cite{Gao2023,Luo2020}, the vast number of parameters usually can not be fulfilled in practice. To solve the model effectively, one approach is to simplify the loss function by eliminating the boundary condition error term. In the study of \cite{Gu2021}, Gu et al. proposed to construct special DNN structures based on boundary conditions. Specifically, for the problem \eqref{01}, a suitable DNN $\phi$ can be designed such that $\phi(x,\theta)=g(x)$ holds on $\partial\Omega$. Then, the boundary condition error in \eqref{02} disappears automatically. The optimization problem \eqref{02} can then be simplified as follows:
\begin{equation}\label{03}
 \min_{\theta}\frac{1}{N_1}\sum_{n=1}^{N_1}\left|\mathcal{D}[\phi(\bx_n^{(1)};\theta),\bx_n^{(1)}]-f(\bx_n^{(1)})\right|^2.
\end{equation}
Compared with \eqref{02}, the simplified minimization \eqref{03} is easier to implement and optimize, obtaining better numerical results in practice.

The optimization \eqref{03} can be viewed as a least squares regression problem, where the target function $f$ is approximated by the special neural network (i.e., PINN) $\phi$. Comparatively, in deep Ritz model, the approximate network $\phi$ is involved in a special energy functional, which is to be minimized. And in WAN model, $\phi$ is trained through a min-max optimization that is derived from the variational form of the PDE. However, for all network-based models, if $\phi$ is moderately deep, training with gradient descent may still suffer from difficulties, such as the gradient vanishing problem, making the optimization stagnated \cite{Rognvaldsson1994,Erhan2009}. 

In \cite{Liu2025}, Liu et al. developed a layer separation (LySep) framework for least squares regression problems with fully connected neural networks (FNNs). Specifically, the output of each layer of the FNN is represented by an auxiliary variable so that the deep network architecture can be split into a series of shallow architectures. These new variables and their constraints are added to the loss function via a weighted quadratic penalty term, and the new loss function only has shallow structures with variables. The LySep model can be solved by specific algorithms using closed-form solutions, and there is no vanishing gradient or explosion gradient problem during the optimization process.

In this work, we develop the LySep models for high-order PDEs. Specifically, we discuss the methods for common second-order linear PDEs (elliptic/parabolic/hyperbolic equations) in detail, where particular LySep models are formulated based on the least squares regression model \eqref{03}. We employ auxiliary variables to represent the output and its derivatives of every layer of $\phi$ and formulate the relations between network parameters. The LySep loss function is then constructed by combining these relations using self-weighted penalty terms. Theoretical analysis shows that the proposed LySep model is mathematically consistent with the original PINN model. 

In the LySep loss function, only the variables of two adjacent layers are coupled. Based on this property, we design specific alternating direction algorithms to solve the optimization, where many variables are updated optimally by closed-form expressions rather than gradient descent. In numerical experiments, LySep performs much more accurately than PINN in solving high-dimensional PDE examples (e.g., 10-dimensional elliptic equations). In spite of the detailed discussion of second-order linear PDEs, we remark that similar approaches can be developed for high-order or nonlinear PDEs, which follow the same strategy but may need more auxiliary variables.

For the sake of brevity, this work discusses only the LySep model based on PINN. We remark that similar models can be established based on other network-based methods, such as the deep Ritz method and WAN. Also, we only discuss the models for second-order linear PDEs as examples, but we do not utilize any special properties of these equations; the strategies could be generalized for high-order or nonlinear PDEs.

The paper is organised as follows: In Section \ref{sec:Elliptic PDE}, we present the LySep model for elliptic PDEs, prove its consistency with the original PINN model, and design the corresponding algorithm. Section \ref{sec:time-dependent PDEs} generalizes the LySep model to parabolic and hyperbolic PDEs. In Section \ref{sec_experiments}, numerical experiments validate the effectiveness of LySep models in solving high-dimensional PDEs and provide numerical verification of the theoretical results. Finally, Section \ref{sec:concludsion} provides conclusions and discusses potential directions for future research work.

\section{Elliptic Partial Differential Equations}\label{sec:Elliptic PDE}

In this section, we consider the following elliptic equation in the $d$-dimensional unit ball $\Omega=\{\bx \in\mathbb{R}^d,\|\bx\|_2\leq 1\}$ with homogeneous Dirichlet boundary condition:
\begin{equation}\label{eq:Elliptic equation}
  \begin{cases}
    \nabla\cdot(c(\bx)\nabla u(\bx))=f(\bx), & \text{for}~ \bx\in\Omega, \\
    u(\bx)= 0 ,            & \text{for}~ \bx\in\partial\Omega,
  \end{cases}
\end{equation}
where $u: \Omega\rightarrow\mathbb{R}$ is the unknown solution; $f:\Omega\rightarrow \mathbb{R}$ is a given data function; $c(\bx)$ is a continuous function with $c(\bx)>\delta$ for some $\delta>0$. 

In the following, we first introduce the PINN model for elliptical PDEs. Next, we propose a LySep model and rigorously prove its consistency with the original PINN model. For the brevity of mathematical deduction and notations, we only formulate the PINN and LySep models with three-layer neural networks. For deeper networks, the LySep model can be derived similarly, but its formulation is more complicated.

\subsection{Fully Connected Neural Network}
We first introduce FNN, one of the most widely used neural networks. Mathematically, let $\sigma(z)$ be an activation function that is applied element-by-element to a vector $z$ to obtain another vector of the same size. 
Define $L\in\mathbb{N}^{+}$ and $M\in\mathbb{N}^{+}$ as the depth and width of the network, respectively. Then FNN $\phi:\mathbb{R}^d\rightarrow\mathbb{R}$ is a combination of $L-1$ simple nonlinear functions given by
\begin{equation}\label{eq:FNN}
\phi(z;\theta)=W_L\sigma(W_{L-1}\sigma(\dots\sigma(W_1z+b_1)\dots)+b_{L-1})+b_L,
\end{equation}
where $W_1\in \mathbb{R}^{M\times d}$, $W_2,\dots,W_{L-1}\in \mathbb{R}^{M\times M}$, $W_L\in \mathbb{R}^{1\times M}$ are weights; $b_1,\dots,b_{L-1}\in \mathbb{R}^{M\times 1}$, $b_L\in \mathbb{R}$ are biases; $\theta=\{W_l,b_l\}_{l=1}^L$ is the set of all free parameters. The number of free parameters is $|\theta| = O(M^2L+Md)$. Due to its special nonlinear structure, the number of free parameters in the FNN increases mildly with dimension, which makes it possible to achieve high-dimensional approximations within a limited storage space. On the other hand, for special functions (e.g., continuous functions \cite{Zhang2022}, H{\"o}lder functions \cite{Jiao2023,Shen2021_1,Shen2021_2}, and Barron functions \cite{Barron1992,Barron1993,Caragea2023,E2020,E2022,Klusowski2018,Siegel2020_1,Siegel2022}), the approximation error rate of neural networks is independent of the dimensionality, thus overcoming or alleviating the curse of dimensionality. In contrast, the degrees of freedom of linear structures such as finite elements and tensor product polynomials grow exponentially as $d$ increases. Therefore, FNNs are more frequently used in high-dimensional approximations.

\subsection{PINN Model for the Elliptic Equations}
In the PINN model for \eqref{eq:Elliptic equation}, one can define a three-layer neural network $\psi_{\e}(\bx;\theta):\mathbb{R}^d\rightarrow\mathbb{R}$,
\begin{equation}\label{eq:FNN-E}
    \psi_{\e}(\bx;\theta)= (\|\bx\|_2^2-1)\phi(\bx;\theta),
\end{equation}
to approximate the PDE solution $u$, where $\phi$ is an FNN defined by \eqref{eq:FNN} with $L=3$ and $z=\bx$; $\|\cdot\|_2$ is the $2$-norm of a vector. Note that $\psi_{\e}(\bx;\theta)$ always satisfies the homogeneous Dirichlet boundary condition. Then, applying the differential operator to $\psi_{\e}(\bx;\theta)$ leads to a PINN $\Phi_{\e}(\bx,\theta)$; namely,
\begin{equation}\label{eq:PINN-E}
\Phi_{\e}(\bx;\theta)=\nabla\cdot(c(\bx)\nabla \psi_{\e}(\bx;\theta)).
\end{equation}
For elliptic equations \eqref{eq:Elliptic equation}, we let $\{\bx_n\}_{n=1}^N$ be a set of feature points in $\Omega$ and $y_n = f(\bx_n)$ for every $n$, then $\{\bx_n,y_n\}_{n=1}^N$ forms a training dataset. 
The least squares PINN model for the problem \eqref{eq:Elliptic equation} is
\begin{equation}\label{eq:PINN-OR-e}
    \min_{\theta}\frac{1}{N}\sum_{n=1}^{N}|\Phi_{\e}(\bx_n;\theta)-y_n|^2.
\end{equation}
For simplicity, we view $\{\bx_n\}_{n=1}^N$ and $\{y_n\}_{n=1}^N$ as column vectors, and use the notations
\begin{gather*}
    X:=\left[\bx_1~\bx_2~\cdots ~\bx_N\right]\in \mathbb{R}^{d\times N},\\
    Y:=\left[y_1~ y_2~\cdots~ y_N\right]\in \mathbb{R}^{1\times N},\\
    \hat{X}:= \left[\|\bx_1\|_2^2~~\|\bx_2\|_2^2~~\cdots~~\|\bx_N\|_2^2\right]\in \mathbb{R}^{1\times N}.
\end{gather*}
Then, \eqref{eq:PINN-OR-e} can be rewritten as follows:
\begin{multline}\label{eq:J-elliptic}
    \min_{W_l,b_l}\Je:=\frac{1}{N}\Big\|\sum_{i=1}^d\pxi\Big(c(X)*\pxi\Big((\hat{X}-1)*(W_3\sigma(W_2\sigma(W_1X+b_1\Bt)
    +b_2\Bt)+b_3\Bt)\Big)\Big)-Y\Big\|_\tF^2,
\end{multline}
where $\Bone$ is the all-one column vector of size $N$ and $\|\cdot\|_\tF$ is the Frobenius norm. 

Existing optimization algorithms (e.g., gradient descent) exhibit notable challenges in solving \eqref{eq:J-elliptic}, especially when the network is very deep. Specifically, due to the non-convexity of the loss function $\Je$, gradient descent tends to converge to bad local minima or suffer from vanishing gradients, resulting in poor convergence performance.

\subsection{Layer Separation with Auxiliary Variables}

In \cite{Liu2025}, the authors proposed a self-adaptive weighted LySep model for the conventional least squares FNNs model by introducing auxiliary variables, which decomposes the deep network into a sequence of shallow networks. Now, we develop the LySep model for second-order PDEs. First, we introduce the auxiliary variables
\begin{equation}\label{eq:var_a}
a_l=\begin{cases}
W_1X+b_1\Bone^\top,\quad &l=1,\\
W_2\sigma(a_1)+b_2\Bone^\top,\quad &l=2.
\end{cases}
\end{equation}
Then, we consider the representation of $\pxi \psi_{\e}$ and $\pxii \psi_{\e}$ by auxiliary variables. Specifically, we introduce 
\begin{equation}\label{eq:var_d_q}
d_{li}=\pxi a_l ~~\text{and}~~ q_{i}=\pxi d_{2i},
\end{equation}
for $l=1,2$ and $i=1,\ldots,d$. 
Using the expression \eqref{eq:var_a} and chain rule, it follows \eqref{eq:var_d_q} that 
\begin{equation}\label{eq:var_d}
d_{li}=\begin{cases}
W_1(:,i)\Bone^\top,\quad &l=1,\\
W_2\left(\sigma'(a_1)*d_{1i}\right),\quad &l=2,
\end{cases}
\end{equation}
and
\begin{equation}\label{eq:var_q}
q_i=W_2\left(\sigma''(a_1)*d_{1i}*d_{1i}\right),
\end{equation}
where $W_1(:,i)$ means the $i$-th column of $W_1$.

Next, we reformulate the deep loss function $\Je$ as a shallower one by embedding the auxiliary variables $a_l$, $d_{li}$ and $q_{i}$, reducing the non-convexity and making it easier to optimize. Using one-step of chain-rule for $\Phi_{\e}(X;\theta)$, we obtain
\begin{equation}\label{eq:Phi}
\begin{split}
    \Phi_{\e}(X;\theta)&=\sum_{i=1}^d\pxi(c(X)*\pxi((\hat{X}-1)*\phi(X;\theta))) \\
    &= K^{\mathrm{\e}}*\phi(X;\theta)
    +\sum_{i=1}^dK^{\e}_i*\pxi\phi(X;\theta)
    +\hat K^{\e}*\sum_{i=1}^d\pxii\phi(X;\theta),
\end{split}
\end{equation}
where $$K^{\mathrm{\e}}=2\Big(dc(X)+\sum_{i=1}^dX(i,:)*\pxi c(X)\Big),\quad \hat K^{\e}=c(X)*(\hat{X}-1),$$
and 
$$K^{\e}_i=4X(i,:)*c(X)+\pxi c(X)*(\hat{X}-1),$$
for $i=1,\ldots,d$. Then, using the chain rule again and performing variable substitution, we get
\begin{equation}\label{eq:pxi_phi}
\begin{split}
&\pxi\phi(X;\theta)=\pxi(W_3\sigma(W_2\sigma(W_1X+b_1\Bt)+b_2\Bt)+b_3\Bt)\\
=&W_3\Big(\sigma'(W_2\sigma(W_1X+b_1\Bt)+b_2\Bt)*(W_2(\sigma'(W_1X+b_1\Bt)*W_1(:,i)\Bt))\Big)\\
=&W_3(\sigma'(a_2)*d_{2i}),
\end{split}
\end{equation}
and
\begin{equation}\label{eq:pxii_phi}
\begin{split}
&\pxii\phi(X;\theta)=\pxii(W_3\sigma(W_2\sigma(W_1X+b_1\Bt)+b_2\Bt)+b_3\Bt)\\
=&W_3\Big(\sigma''(W_2\sigma(W_1X+b_1\Bt)+b_2\Bt)*\pxi(W_2\sigma(W_1X+b_1\Bt)+b_2\Bt)\\
&*\pxi(W_2\sigma(W_1X+b_1\Bt)+b_2\Bt)+\sigma'(W_2\sigma(W_1X+b_1\Bt)+b_2\Bt)\\
&*(W_2(\sigma''(W_1X+b_1\Bt)*W_1(:,i)\Bt*W_1(:,i)\Bt))\Big)\\
=&W_3(\sigma''(a_2)*d_{2i}*d_{2i}+\sigma'(a_2)*q_{i}),
\end{split}
\end{equation}
for $i=1,\ldots,d$. By performing a variable substitution on $\phi(X;\theta)$ and substituting it along with \eqref{eq:pxi_phi} and \eqref{eq:pxii_phi} into \eqref{eq:Phi}, the loss function $\Je$ in \eqref{eq:J-elliptic} becomes the loss term:
\begin{equation}\label{loss_term}
\begin{split}
\frac{1}{N}\Big\|&K^{\e}*(W_3\sigma(a_2)+b_3\Bt)+\sum_{i=1}^dK^{\e}_i*(W_3(\sigma'(a_2)*d_{2i}))\\
&+\hat K^{\e}*\sum_{i=1}^dW_3(\sigma''(a_2)*d_{2i}*d_{2i}+\sigma'(a_2)*q_{i})-Y\Big\|_\tF^2.
\end{split}
\end{equation}

Next, we add these new variables and their constraints to \eqref{loss_term} via least squares penalty terms to construct the total loss function. To simplify the notations, we define the following matrix norm for any $A\in\mathbb{R}^{M\times N}$:
$$\|A\|_\mathcal{D}=\left(\sum_{n=1}^N\mathcal{D}_n^2\|A(:,n)\|_2^2\right)^\frac{1}{2},$$
where $\mathcal{D}=\mathrm{diag}(\mathcal{D}_1,\dots,\mathcal{D}_N)$. Note that $\mathcal{D}$ assigns its diagonals as weights to each column of the matrix $A$ in the computation of the norm. And clearly, $\|A\|_\mathcal{D}=\|A\|_\tF$ if $\mathcal{D}$ is the identity matrix. By coupling all auxiliary variables into least squares penalty terms, we obtain the following type of new loss:
\begin{equation*}
\begin{split}
\Jse:=\frac{1}{N}\Bigg[\Big\|&K^{\e}*(W_3\sigma(a_2)+b_3\Bt)+\sum_{i=1}^dK^{\e}_i*(W_3(\sigma'(a_2)*d_{2i}))\\
&+\hat K^{\e}*\sum_{i=1}^dW_3(\sigma''(a_2)*d_{2i}*d_{2i}+\sigma'(a_2)*q_{i})-Y\Big\|_\tF^2\\
&+\omega_{a_1}\|W_1X+b_1\Bone^\top-a_1\|_{\mathcal{D}_{a_1}}^2
+\omega_{a_2}\|W_2\sigma(a_1)+b_2\Bone^\top-a_2\|_{\mathcal{D}_{a_2}}^2\\
&+\sum_{i=1}^d\Big(\omega_{d_{1i}}\|W_1(:,i)\Bt-d_{1i}\|_{\mathcal{D}_{d_{1i}}}^2+\omega_{d_{2i}}\|W_2(\sigma'(a_1)*d_{1i})-d_{2i}\|_{\mathcal{D}_{d_{2i}}}^2\\
&+\omega_{q_i}\|W_2(\sigma''(a_1)*d_{1i}*d_{1i})-q_{i}\|_{\mathcal{D}_{q_i}}^2\Big)\Bigg],
\end{split}
\end{equation*}
where $\{\omega\}$ (indices omitted) are specific weighting terms that balance these penalty terms, and $\{\mathcal{D}\}$ are specific diagonal matrices.

A natural way is to set above $\{\omega\}$ and $\{\mathcal{D}\}$ as constants. However, the LySep model with constant weights is usually inconsistent with the original least squares model. Namely, decreasing the loss $\Jse$ can not guarantee that the original loss $\Je$ will also decrease. Therefore, it is necessary to construct self-adaptive $\{\omega\}$ and $\{\mathcal{D}\}$ to ensure the consistency between $\Jse$ and $\Je$. That is, we need to construct specific relation between $\{\omega\}$ and $\{\mathcal{D}\}$ and the variables $\{W_l,b_l,a_l,d_{li},q_{i}\}$.

For elliptic equations, we propose the LySep loss function with auxiliary variables as
\begin{multline}\label{eq:Js-elliptic}
    \Jse= \frac{1}{N}\Bigg[\Big\|K^{\e}*(W_3\sigma(a_2)+b_3\Bt)+\sum_{i=1}^dK^{\e}_i*(W_3(\sigma'(a_2)*d_{2i}))
    +\hat K^{\e}*\sum_{i=1}^dW_3(\sigma''(a_2)*d_{2i}*d_{2i}+\sigma'(a_2)*q_{i})-Y\Big\|_\tF^2\\
    +\omega_{a_1}^1\|W_1X+b_1\Bone^\top-a_1\|_{\mathcal{D}_{a_1}^1}^2+\sum_{k=2}^3\sum_{i=1}^d\omega_{a_1}^{ki}\|W_1X+b_1\Bone^\top-a_1\|_{\mathcal{D}_{a_1}^{ki}}^2\\
    +\omega_{a_2}^1\|W_2\sigma(a_1)+b_2\Bone^\top-a_2\|_{\mathcal{D}_{a_2}^1}^2+\sum_{i=1}^d\omega_{a_2}^{2i}\|W_2\sigma(a_1)+b_2\Bone^\top-a_2\|_{\mathcal{D}_{a_2}^{2i}}^2\\
   +\sum_{k=1}^2\sum_{i=1}^d\Big(\omega_{d_{1i}}^{k}\|W_1(:,i)\Bt-d_{1i}\|_{\mathcal{D}_{d_{1i}}^{k}}^2+\omega_{d_{2i}}^{k}\|W_2(\sigma'(a_1)*d_{1i})-d_{2i}\|_{\mathcal{D}_{d_{2i}}^{k}}^2\Big)\\
   +\sum_{i=1}^d\omega_{q_i}\|W_2(\sigma''(a_1)*d_{1i}*d_{1i})-q_{i}\|_\tF^2\Bigg],
\end{multline}
where the specific expressions for the weight terms are as follows
\begin{align*}
&\omega_{a_1}^1=\omega_{d_{1i}}^{2}=\|W_3\|_\tF^2\|W_2\|_\tF^2;\quad \omega_{a_1}^{2i}=\omega_{a_2}^{2i}=\omega_{a_1}^1\|W_1(:,i)\|_\tF^2;\quad
\omega_{a_1}^{3i}=\omega_{a_1}^{2i}\|W_2\|_\tF^2;\\ 
&\omega_{a_2}^1=\omega_{d_{2i}}^{1}=\|W_3\|_\tF^2;\quad
\omega_{d_{1i}}^{1}=|\hat K^{\e}|_\infty^2(\omega_{a_1}^{2i}+\omega_{a_1}^{3i});\quad
\omega_{d_{2i}}^{2}=|\hat K^{\e}|_\infty^2\omega_{a_1}^{2i};\quad
\omega_{q_i}=|\hat K^{\e}|_\infty^2\omega_{a_2}^{1},
\end{align*}
and the specific expressions for the diagonal matrices are as follows
\begin{align*}
&\mathcal{D}_{a_1}^1(n,n)=\Big(\mathcal{D}_{a_2}^1(n,n)^2+\sum_{i=1}^d\Big(|K^{\e}_i|_{\infty}^2\|d_{1i}[n]\|_\tF^2+\mathcal{D}_{a_1}^{2i}(n,n)^2\|d_{2i}[n]\|_\tF^2\Big)\Big)^{\frac 12};\\ 
&\mathcal{D}_{a_1}^{2i}(n,n)=|\hat K^{\e}|_\infty\|d_{1i}[n]\|_\tF;\quad
\mathcal{D}_{a_1}^{3i}(n,n)=\Big(\mathcal{D}_{a_1}^{2i}(n,n)^2+\mathcal{D}_{a_2}^{2i}(n,n)^2\Big)^{\frac 12}\\
&\mathcal{D}_{a_2}^1(n,n)=\Big(|K^{\e}|_{\infty}^2+\sum_{i=1}^d(|K^{\e}_i|_{\infty}^2\|d_{2i}[n]\|_\tF^2+|\hat K^{\e}|_{\infty}^2\|q_{i}[n]\|_\tF^2)\Big)^{\frac 12};\\ 
&\mathcal{D}_{a_2}^{2i}(n,n)=|\hat K^{\e}|_\infty\|d_{2i}[n]\|_\tF;\quad
\mathcal{D}_{d_{1i}}^{1}=\mathcal{D}_{d_{2i}}^{2}=I;\\ 
&\mathcal{D}_{d_{1i}}^{2}(n,n)=\Big(|K^{\e}_i|_{\infty}^2+\mathcal{D}_{a_1}^{3i}(n,n)^2\Big)^{\frac 12};\quad
\mathcal{D}_{d_{2i}}^{1}(n,n)=\Big(|K^{\e}_i|_{\infty}^2+\mathcal{D}_{a_2}^{2i}(n,n)^2\Big)^{\frac 12},
\end{align*}
for $i=1,\ldots,d$ and $n=1,\ldots,N$.

Then, the corresponding LySep model is given by
\begin{equation}\label{model_LySep}
    \min_{W_l,b_l,a_l,d_{li},q_{i}}\Jse.
\end{equation}

For $\Jse$, a key theoretical issue is its consistency relationship with $\Je$. Specifically, we need to verify that when $\Jse$ decreases during the training process of the LySep model, $\Je$ maintains the same decreasing trend. This condition is important for the effectiveness of the LySep model. In the following part, we establish the consistency between $\Jse$ and $\Je$, whose proof is presented in \ref{proof}.
\begin{thm}\label{thm1}
Suppose $\sigma$, $\sigma'$ and $\sigma''$ are Lipschitz continuous and bounded functions, i.e., $|\sigma(z_1)-\sigma(z_2)|\leq C_{\sigma}|z_1-z_2|$, $|\sigma'(z_1)-\sigma'(z_2)|\leq C_{\sigma'}|z_1-z_2|$, $|\sigma''(z_1)-\sigma''(z_2)|\leq C_{\sigma''}|z_1-z_2|$, $|\sigma(z_1)|\leq B_{\sigma}$, $|\sigma'(z_1)|\leq B_{\sigma'}$, and $|\sigma''(z_1)|\leq B_{\sigma''}$, for some $C_{\sigma}, C_{\sigma'}, C_{\sigma''}, B_{\sigma'}, B_{\sigma''}>0$ and any $z_1,z_2\in\mathbb{R}$. Given $\Je$ and $\Jse$ defined in \eqref{eq:J-elliptic} and \eqref{eq:Js-elliptic}, respectively. Then, for all $\{W_l,b_l\}_{l=1}^3$, $\{a_1,a_2\}$, $\{d_{li}\}_{i=1,\ldots,d}^{l=1,2}$, and $\{q_{i}\}_{i=1,\ldots,d}$, it holds that
\begin{equation}\label{eq:thm1-01}
    \Je\leq 2(d+1)\cdot C\cdot\Jse, 
\end{equation}
where $C=\max\{1,2C_{\sigma}^2,2C_{\sigma}^4,5C_1^2,14C_2^2\}$, $C_1=\max\{B_{\sigma'}\cdot\max\{1,B_{\sigma'}\},C_{\sigma'}\cdot\max\{1,B_{\sigma'},C_{\sigma}\}\}$, and $C_2=\max\{B_{\sigma''}\cdot\max\{1,B_{\sigma'},C_{\sigma'},B_{\sigma'}^2,B_{\sigma'}C_{\sigma'}\},\max\{1,C_{\sigma'}\}\cdot\max\{C_{\sigma'},B_{\sigma'}C_{\sigma''}\}\}$.
\end{thm}

\begin{thm}\label{thm2}
For all $\{W_l,b_l\}_{l=1}^3$, there exists $\{a_1,a_2\}$, $\{d_{li}\}_{i=1,\ldots,d}^{l=1,2}$, and $\{q_{i}\}_{i=1,\ldots,d}$, such that
\begin{equation}\label{eq:thm1-02}
\Jse\leq\Je.
\end{equation}
\end{thm}

The above theoretical results show that the LySep model is mathematically consistent with the original least squares PINN model. Specifically, optimizing the LySep loss function $\Jse$ can ensure the simultaneous decrease of the mean square loss $\Je$. Despite the consistency between $\Jse$ and $\Je$ in theory, the difficulty of their optimization in practice varies greatly: the LySep loss function $\Jse$ is a combination of several squared terms with shallow architectures (i.e., there is at most one hidden layer in every squared term), which can be optimized by specific alternating direction algorithms efficiently (see Section \ref{Imp_e}); the mean square loss $\Je$ has a deeper architecture, making common optimization strategies inefficient.
 
\subsection{Implementation}\label{Imp_e}
In this section, we develop an alternating direction algorithm to minimize $\Jse$. Thanks to the shallow architecture, $\Jse$ is convex with respect to many variables such as $W_3,b_1,b_2$, etc., each of which can be computed in closed-form in the minimization. Therefore, in every iteration, we alternatively update all variables, and some variables are updated by solving least squares linear systems. Specifically, we update $W_3$ by the linear system
\begin{equation}\label{eq:GD_W3}
W_3\begin{bmatrix}
\sqrt{\lambda}I~&~B
\end{bmatrix}
=\begin{bmatrix}
O~&~Y-K^{\e}*b_3\Bt
\end{bmatrix},
\end{equation}
where
\begin{align*}
B=&\sigma(a_2)\diag[K^{\e}]+\sum_{i=1}^d\Big(D_{2i}\diag[K_i^{\e}]+(Q_{2i}+\hat{Q}_{2i})\diag[\hat{K}^{\e}]\Big)\Big),\\
\lambda=&\omega_1\|W_1A_1-P_1\|_{\mathcal{D}_{a_1}^{1}}^2+\sum_{i=1}^d\Big(\omega_{2}^i\|W_1A_1-P_1\|_{\mathcal{D}_{a_1}^{2i}}^2+\omega_{3}^i\|W_1A_1-P_1\|_{\mathcal{D}_{a_1}^{3i}}^2\Big)\\
&+\|W_2A_2-P_2\|_{\mathcal{D}_{a_2}^{1}}^2+\sum_{i=1}^d\Big(\omega_{2}^i\|W_2A_2-P_2\|_{\mathcal{D}_{a_2}^{2i}}^2+\omega_1\|W_1(:,i)\Bt-d_{1i}\|_{\mathcal{D}_{d_{1i}}^{2}}^2\\
&+(\omega_{2}^i+\omega_3^i)\|W_1(:,i)\Bt-d_{1i}\|_{\mathcal{D}_{d_{1i}}^{1}}^2+\|W_2D_{1i}-d_{2i}\|_{\mathcal{D}_{d_{2i}}^{1}}^2\\
&+|\hat{K}^{\e}|_\infty^2\omega_{2}^i\|W_2D_{1i}-d_{2i}\|_{\mathcal{D}_{d_{2i}}^{2}}^2
+|\hat{K}^{\e}|_\infty^2\|W_2Q_{1i}-q_i\|_\tF^2\Big), 
\end{align*}
and 
\begin{align*}
\omega_1=\|W_2\|_\tF^2,\quad \omega_2^i=\|W_2\|_\tF^2\|W_1(:,i)\|_\tF^2,\quad \omega_3^i=\|W_2\|_\tF^4\|W_1(:,i)\|_\tF^2,
\end{align*}
with
\begin{align*}
A_1=X,\quad A_l=\sigma(a_{l-1}),\quad P_l=a_l-b_l\Bt,
\end{align*}
for $l=2,3$, and
\begin{align*}
D_{li}=\sigma'(a_l)*d_{li},\quad Q_{li}=\sigma''(a_l)*d_{li}*d_{li},\quad
\hat{Q}_{2i}=\sigma'(a_2)*q_{i},
\end{align*}
for $l=1,2$ and $i=1,\ldots,d$.

Similarly, $b_1$, $b_2$ and $b_3$ are updated by the linear systems
\begin{equation}\label{eq:GD_b1}
b_1\Bt\begin{bmatrix}
\mathcal{D}_{a_1}^1~&~\mathcal{D}_{a_1}^2~&~\mathcal{D}_{a_1}^3
\end{bmatrix}
=(a_1-W_1A_1)
\begin{bmatrix}
\mathcal{D}_{a_1}^1~&~\mathcal{D}_{a_1}^2~&~\mathcal{D}_{a_1}^3
\end{bmatrix},
\end{equation}
\begin{equation}\label{eq:GD_b2}
b_2\Bt\begin{bmatrix}
\mathcal{D}_{a_2}^1~&~\mathcal{D}_{a_2}^2
\end{bmatrix}
=(a_2-W_2A_2)
\begin{bmatrix}
\mathcal{D}_{a_2}^1~&~\mathcal{D}_{a_2}^2
\end{bmatrix},
\end{equation}
and
\begin{equation}\label{eq:GD_b3}
K^{\e}*b_3\Bt=Y-W_3B,
\end{equation}
respectively, where
\begin{align*}
&\mathcal{D}_{a_1}^2=\begin{bmatrix}
    \|W_1(:,1)\|_\tF\mathcal{D}_{a_1}^{21}& \cdots & \|W_1(:,d)\|_\tF\mathcal{D}_{a_1}^{2d}
\end{bmatrix},\\
& \mathcal{D}_{a_2}^2=\|W_2\|_\tF\begin{bmatrix}
    \|W_1(:,1)\|_\tF\mathcal{D}_{a_2}^{21}& \cdots & \|W_1(:,d)\|_\tF\mathcal{D}_{a_2}^{2d}
\end{bmatrix},\\
&\mathcal{D}_{a_1}^3=\|W_2\|_\tF\begin{bmatrix}
    \|W_1(:,1)\|_\tF\mathcal{D}_{a_1}^{31}& \cdots & \|W_1(:,d)\|_\tF\mathcal{D}_{a_1}^{3d}
\end{bmatrix}.
\end{align*}

The variables $\{W_l,a_l\}_{l=1}^2$, and $\{d_{li}\}_{i=1}^d$ involved in $\Jse$ are non-convex, but they are involved in shallow architectures in the loss function, which eases the optimization to some extent. Here, we update these variables by gradient descent.

To sum up, we present the algorithm that solves the optimization \eqref{model_LySep} (see Algorithm \ref{alg:01}).
\begin{algorithm}
\DontPrintSemicolon
\SetAlCapFnt{\fontsize{11pt}{18pt}}
\fontsize{10pt}{12pt}\selectfont
\KwIn{data $X$, $Y$; number of iterations $N_k$; learning rate $\tau$}
\KwOut{a feasible solution $\{W_l,b_l,a_l,d_{li},q_{i}\}$.}
\Begin{initialize $\{W_l,b_l\}_{l=1}^3$, $\{a_l,\{d_{li}\}_{i=1}^d\}_{l=1}^2$, and $\{q_{i}\}_{i=1}^d$\;
\For{$k = 0,\cdots,N_k-1$}{
\For{$i = 1,\cdots,d$}{
$W_1(:,i)\leftarrow W_1(:,i)-\tau\nabla_{W_1(:,i)}\Jse$
}
Solve \eqref{eq:GD_b1} for $b_1$\;
$a_1\leftarrow a_1-\tau\nabla_{a_1}\Jse$\;
\For{$i = 1,\cdots,d$}{
$d_{1i}\leftarrow d_{1i}-\tau\nabla_{d_{1i}}\Jse$
}
$W_2\leftarrow W_2-\tau\nabla_{W_2}\Jse$\;
Solve \eqref{eq:GD_b2} for $b_2$\;
$a_2\leftarrow a_2-\tau\nabla_{a_2}\Jse$\;
\For{$i = 1,\cdots,d$}{
$d_{2i}\leftarrow d_{2i}-\tau\nabla_{d_{2i}}\Jse$
}
\For{$i = 1,\cdots,d$}{
$q_{i}\leftarrow q_{i}-\tau\nabla_{q_{i}}\Jse$
}
Solve \eqref{eq:GD_W3} for $W_3$\;
Solve \eqref{eq:GD_b3} for $b_3$\;
}
return $\{W_l,b_l\}_{l=1}^3$, $\{a_l,\{d_{li}\}_{i=1}^d\}_{l=1}^2$, and $\{q_{i}\}_{i=1}^d$.
}
\caption{Solve $\min_{W_l,b_l,a_l,d_{li},q_{i}} \Jse$}\label{alg:01}
\end{algorithm}

\section{Time-dependent Partial differential equations}\label{sec:time-dependent PDEs}

In this section, we will consider time-dependent PDEs with homogeneous boundary and initial conditions, including the parabolic equation
\begin{equation}\label{eq:Parabolic equation}
    \begin{cases}
       \frac{\partial u}{\partial t}(t,\bx)=\nabla\cdot(c(\bx)\nabla u(t,\bx))+Q(t,\bx),&~\text{for} ~(t,\bx)\in (0,T]\times\Omega,\\
        u(t,\bx)=0,&~\text{for} ~(t,\bx)\in \{0\}\times\Omega\cup(0,T]\times\partial\Omega,
        \end{cases}
\end{equation}
and the hyperbolic equation
\begin{equation}\label{eq:Hyperbolic equation}
    \begin{cases}
       \frac{\partial^2 u}{\partial t^2}(t,\bx)=\nabla\cdot(c(\bx)\nabla u(t,\bx))+Q(t,\bx),&~\text{for} ~(t,\bx)\in (0,T]\times\Omega\\
        u(t,\bx)=0,&~\text{for} ~(t,\bx)\in \{0\}\times\Omega\cup(0,T]\times\partial\Omega,\\
        \frac{\partial u}{\partial t}(t,\bx)=0,&~\text{for} ~(t,\bx)\in\{0\}\times \Omega,\\
        \end{cases}
\end{equation}
where $u: [0,T]\times\Omega\rightarrow\mathbb{R}$ is the unknown solution; $Q:[0,T]\times\Omega\rightarrow \mathbb{R}$ is a given data function; $c(\bx)$ is a continuous function with $c(\bx)>\delta$ for some $\delta>0$; $\Omega$ is a $d$-dimensional unit ball and $T>0$.

Based on a similar treatment as in Section \ref{sec:Elliptic PDE}, we first introduce PINN models for parabolic and hyperbolic PDEs. Next, we propose the LySep models and prove their consistency with the original PINN models. For the simplicity of mathematical derivation, we still take three-layer neural networks for discussion.

\subsection{PINN Models for Time-dependent Equations}
\subsubsection{PINN model for the parabolic equations}
We introduce the PINN model for \eqref{eq:Parabolic equation}. Let $\psi_{\p}(t,\bx;\theta):\mathbb{R}\times\mathbb{R}^{d}\rightarrow\mathbb{R}$ be the following three-layer neural network
\begin{equation}\label{eq:FNN-P}
    \psi_{\p}(t,\bx;\theta)= t\Big(\sum_{i=1}^d|x_i|^2-1\Big)\phi(t,\bx;\theta),\\
\end{equation}
where $\phi$ is an FNN defined by \eqref{eq:FNN} with $L=3$ and $z=[t~ \bx^\top]^\top$. Note that $\psi_{\p}$ always satisfies the homogeneous boundary and initial conditions and is used to approximate the PDE solution $u(t,\bx)$. Then, applying $\frac{\partial}{\partial t}-\nabla\cdot c(\bx)\nabla$ to $\psi_{\p}(t,\bx;\theta)$ leads to a PINN $\Psi_{\p}(t,\bx;\theta)$; namely,
\begin{equation}\label{eq:PINN-P}
    \Psi_{\p}(t,\bx;\theta)= \frac{\partial\psi_{\p}}{\partial t}(t,\bx;\theta)-\nabla\cdot(c(\bx)\nabla\psi_{\p}(t,\bx;\theta)).
\end{equation}
Let $\{(t_n,\bx_n)\}_{n=1}^N$ be a set of feature points in $[0,T]\times\Omega$ and $y_n = Q(t_n,\bx_n)$ for every $n$, then $\{(t_n,\bx_n,y_n)\}_{n=1}^N$ forms a training dateset. The least squares PINN model for the problem \eqref{eq:Parabolic equation} is given by
\begin{equation}\label{eq:PINN-OR-p}
    \min_{\theta}\frac{1}{N}\sum_{n=1}^N\Big|\Psi_{\p}(t_n,\bx_n;\theta)-y_n\Big|^2.
\end{equation}
For simplicity, we follow the notations of $X$, $Y$ and $\hat{X}$ in Section \ref{sec:Elliptic PDE}, and define
\begin{gather*}
    T_0=\left[~t_1~t_2~\cdots ~t_N~\right]\in \mathbb{R}^{1\times N},\quad
    X_T=\left[~T_0;~X~\right]\in \mathbb{R}^{(d+1)\times N}.
\end{gather*}
Then, \eqref{eq:PINN-OR-p} can be rewritten as follows:
\begin{gather}\label{eq:J-parabolic}
    \min_{W_l,b_l} \Jp:=\frac{1}{N}\Big\|\pt\Big(T_0*(\hat X -1)*(W_3\sigma(W_2\sigma(W_1X_T+b_1\Bt)+b_2\Bt)+b_3\Bt)\Big)\\
    -\sum_{i=1}^d\pxi\Big(c(X)*\pxi\Big(T_0*(\hat X -1)*(W_3\sigma(W_2\sigma(W_1X_T+b_1\Bt)+b_2\Bt)+b_3\Bt)\Big)\Big)-Y\Big\|_\tF^2.\nonumber
\end{gather}

\subsubsection{PINN model for the hyperbolic equations}
We introduce the PINN model for \eqref{eq:Hyperbolic equation}. Similarly, the PDE approximate solution can be chosen as a three-layer neural network $\psi_{\h}(t,\bx;\theta):\mathbb{R}\times\mathbb{R}^{d}\rightarrow\mathbb{R}$,
\begin{equation}\label{eq:FNN-H}
    \psi_{\h}(t,\bx;\theta)= t^2\Big(\sum_{i=1}^d|x_i|^2-1\Big)\phi(t,\bx;\theta),
\end{equation}
where $\phi$ is an FNN defined by \eqref{eq:FNN} with $L=3$ and $z=[t ~\bx^\top]^\top$. Note that $\psi_{\h}$ always satisfies the homogeneous boundary and initial conditions. Then, applying $\frac{\partial^2}{\partial t^2}-\nabla\cdot c(\bx)\nabla$ to $\psi_{\h}(t,\bx;\theta)$ leads to a PINN $\Psi_{\h}$; namely,
\begin{equation}\label{eq:PINN-H}
    \Psi_{\h}(t,\bx;\theta)= \frac{\partial^2\psi_{\h}}{\partial t^2}(t,\bx;\theta)-\nabla\cdot(c(\bx)\nabla \psi_{\h}(t,\bx;\theta)).
\end{equation}
We can obtain the least squares PINN model for the problem \eqref{eq:Hyperbolic equation}:
\begin{equation}\label{eq:PINN-OR-h}
    \min_{\theta}\frac{1}{N}\sum_{n=1}^N\Big|\Psi_{\h}(t_n,\bx_n;\theta)-y_n\Big|^2.
\end{equation}
For simplicity, we also use the notations of $X$, $Y$, $\hat{X}$, $T_0$, and $X_T$, and define
\begin{gather*}
    \hat{T}_0=\left[~t_1^2~t_2^2~\cdots ~t_N^2~\right]\in \mathbb{R}^{1\times N}.
\end{gather*} Then, \eqref{eq:PINN-OR-h} can be rewritten as follows:
\begin{gather}\label{eq:J-hyberbolic}
    \min_{W_l,b_l} \Jh:=\frac{1}{N}\Big\|\ptt\Big(T_0*(\hat X -1)*(W_3\sigma(W_2\sigma(W_1X_T+b_1\Bt)+b_2\Bt)+b_3\Bt)\Big)\\
    -\sum_{i=1}^d\pxi\Big(c(X)*\pxi(T_0*(\hat X -1)*(W_3\sigma(W_2\sigma(W_1X_T+b_1\Bt)+b_2\Bt)+b_3\Bt))\Big)-Y\Big\|_\tF^2.\nonumber
\end{gather}

\subsection{Layer Separation Model and Implementation}

Similarly to Section \ref{sec:Elliptic PDE}, to solve the optimization \eqref{eq:J-parabolic} and \eqref{eq:J-hyberbolic} effectively, we propose the LySep models by introducing auxiliary variables to represent output and its derivatives of every layer, which reduces the non-convexity and makes it easier to optimize. The auxiliary variables and their constraints are added to this loss function through weighted penalty terms. In this section for time-dependent PDEs, we denote $x_0=t$ and redefine the auxiliary variables by
\begin{equation}
a_l=\begin{cases}
W_1X_T+b_1\Bone^\top,\quad &l=1,\\
W_2\sigma(a_1)+b_2\Bone^\top,\quad &l=2,
\end{cases}
\end{equation}
\begin{equation}
d_{li}=\pxi a_l=\begin{cases}
W_1(:,i+1)\Bone^\top,\quad &l=1,\\
W_2\left(\sigma'(a_1)*d_{1i}\right),\quad &l=2,
\end{cases}
\end{equation}
and
\begin{equation}
q_{i}=W_2\left(\sigma''(a_1)*d_{1i}*d_{1i}\right),
\end{equation}
for $i=0,\ldots,d$.

\subsubsection{Methods for parabolic equations}
Similar to \eqref{eq:Phi}-\eqref{eq:pxii_phi}, we also use chain rule and variable substitution for PINN $\Psi_{\p}(t,\bx;\theta)$, and add these auxiliary variables and their constraints to the loss function via weighted penalty terms. Then, we present the LySep loss function with auxiliary variables for problem \eqref{eq:Parabolic equation}:
\begin{multline}\label{eq:Js-parabolic}
    \Jsp := \frac{1}{N}\Bigg[\Big\|K^{\p}*(W_3\sigma(a_2)+b_3\Bt)+K^{\p}_0*(W_3(\sigma'(a_2)*d_{2,0}))-\sum_{i=1}^dK^{\p}_i*(W_3(\sigma'(a_2)*d_{2i}))\\
-\hat K^{\p}*\sum_{i=1}^dW_3(\sigma''(a_2)*d_{2i}*d_{2i}+\sigma'(a_2)*q_{i})-Y\Big\|_\tF^2+\omega_{a_1}^1\|W_1X_T+b_1\Bone^\top-a_1\|_{\mathcal{D}_{a_1}^1}^2\\
+\sum_{k=2}^3\sum_{i=1}^d\omega_{a_1}^{ki}\|W_1X_T+b_1\Bone^\top-a_1\|_{\mathcal{D}_{a_1}^{ki}}^2+\omega_{a_2}^1\|W_2\sigma(a_1)+b_2\Bone^\top-a_2\|_{\mathcal{D}_{a_2}^1}^2\\
+\sum_{i=1}^d\omega_{a_2}^{2i}\|W_2\sigma(a_1)+b_2\Bone^\top-a_2\|_{\mathcal{D}_{a_2}^{2i}}^2
+\omega_{d_{10}}\|W_1(:,1)\Bt-d_{1,0}\|_\tF^2\\
+\omega_{d_{20}}\|W_2(\sigma'(a_1)*d_{1,0})-d_{2,0}\|_\tF^2
    +\sum_{k=1}^2\sum_{i=1}^d\Big(\omega_{d_{1i}}^{k}\|W_1(:,i+1)\Bt-d_{1i}\|_{\mathcal{D}_{d_{1i}}^{k}}^2\\
    +\omega_{d_{2i}}^{k}\|W_2(\sigma'(a_1)*d_{1i})-d_{2i}\|_{\mathcal{D}_{d_{2i}}^{k}}^2\Big)
   +\sum_{i=1}^d\omega_{q_i}\|W_2(\sigma''(a_1)*d_{1i}*d_{1i})-q_{i}\|_\tF^2\Bigg],
\end{multline}
where 
\begin{align*}
    K^{\p}=(\hat X-1)-2dc(X)*T_0-2T_0*\sum_{i=1}^d\pxi c(X)*X(i,:);\quad K^{\p}_0=T_0*(\hat X-1);\\
    K^{\p}_i= T_0*\left((\hat{X}-1)*\pxi c(X)+4X(i,:)*c(X)\right);\quad \hat K^{\p}=c(X)*T_0*(\hat X-1),
\end{align*}
the specific expressions for the weight terms are as follows
\begin{gather*}
\omega_{a_1}^1=\omega_{d_{1i}}^{2}=\|W_3\|_\tF^2\|W_2\|_\tF^2;\quad
\omega_{a_1}^{2i}=\omega_{a_2}^{2i}=\omega_{a_1}^1\|W_1(:,i+1)\|_\tF^2;\\
\omega_{a_1}^{3i}=\omega_{a_1}^{2i}\|W_2\|_\tF^2;\quad
\omega_{a_2}^1=\omega_{d_{2i}}^{1}=\|W_3\|_\tF^2;\quad
\omega_{d_{10}}=|K^{\p}_0|_\infty^2\omega_{a_1}^{1};\\
\omega_{d_{20}}=|K^{\p}_0|_\infty^2\omega_{a_2}^{1};\quad
\omega_{d_{1i}}^{1}=|\hat K^{\p}|_\infty^2(\omega_{a_1}^{2i}+\omega_{a_1}^{3i});\quad
\omega_{d_{2i}}^{2}=|\hat K^{\p}|_\infty^2\omega_{a_1}^{2i};\quad
\omega_{q_i}=|\hat K^{\p}|_\infty^2\omega_{a_2}^{1},
\end{gather*}
and the specific expressions for the diagonal matrices are as follows
\begin{align*}
&\mathcal{D}_{a_1}^1(n,n)=\Big(\mathcal{D}_{a_2}^{1}(n,n)^2+\sum_{i=0}^d|K^{\p}_i|_{\infty}^2\|d_{1i}[n]\|_\tF^2+\sum_{i=1}^d\mathcal{D}_{a_1}^{2i}(n,n)^2\|d_{2i}[n]\|_\tF^2\Big)^{\frac 12};\\ 
&\mathcal{D}_{a_1}^{2i}(n,n)=|\hat K^{\p}|_\infty\|d_{1i}[n]\|_\tF;\quad \mathcal{D}_{a_1}^{3i}(n,n)=\Big(\mathcal{D}_{a_1}^{2i}(n,n)^2+\mathcal{D}_{a_2}^{2i}(n,n)^2\Big)^{\frac 12}\\
&\mathcal{D}_{a_2}^1(n,n)=\Big(|K^{\p}|_{\infty}^2+\sum_{i=0}^d|K^{\p}_i|_{\infty}^2\|d_{2i}[n]\|_\tF^2+|\hat K^{\p}|_{\infty}^2\sum_{i=1}^d\|q_{i}[n]\|_\tF^2\Big)^{\frac 12};\\ 
&\mathcal{D}_{a_2}^{2i}(n,n)=|\hat K^{\p}|_\infty\|d_{2i}[n]\|_\tF;\quad \mathcal{D}_{d_{1i}}^{1}=\mathcal{D}_{d_{2i}}^{2}=I;\\
&\mathcal{D}_{d_{1i}}^{2}(n,n)=\Big(|K^{\p}_i|_{\infty}^2+\mathcal{D}_{a_1}^{3i}(n,n)^2\Big)^{\frac 12};\quad \mathcal{D}_{d_{2i}}^{1}(n,n)=\Big(|K^{\p}_i|_{\infty}^2+\mathcal{D}_{a_2}^{2i}(n,n)^2\Big)^{\frac 12},
\end{align*}
for $i=1,\ldots,d$ and $n=1,\ldots,N$.

Then, we can obtain the LySep model for the parabolic equation:
\begin{equation}\label{model_LySep2}
    \min_{W_l,b_l,a_l,d_{li},q_{i}}\Jsp.
\end{equation}

For $\Jsp$, its consistency with $\Jp$ is also important. For this, we give the following theoretical results:
\begin{thm}\label{thm3}
Suppose $\sigma$, $\sigma'$ and $\sigma''$ are Lipschitz continuous and bounded functions, i.e., $|\sigma(z_1)-\sigma(z_2)|\leq C_{\sigma}|z_1-z_2|$, $|\sigma'(z_1)-\sigma'(z_2)|\leq C_{\sigma'}|z_1-z_2|$, $|\sigma''(z_1)-\sigma''(z_2)|\leq C_{\sigma''}|z_1-z_2|$, $|\sigma(z_1)|\leq B_{\sigma}$, $|\sigma'(z_1)|\leq B_{\sigma'}$ and $|\sigma''(z_1)|\leq B_{\sigma''}$, for some $C_{\sigma}, C_{\sigma'}, C_{\sigma''}, B_{\sigma'}, B_{\sigma''}>0$ and any $z_1,z_2\in\mathbb{R}$. Given $\Jp$ and $\Jsp$ defined in \eqref{eq:J-parabolic} and \eqref{eq:Js-parabolic}, respectively. Then, for all $\{W_l,b_l\}_{l=1}^3$, $\{a_1,a_2\}$, $\{d_{li}\}_{i=0,\ldots,d}^{l=1,2}$, and $\{q_{i}\}_{i=1,\ldots,d}$, it holds that
\begin{equation}\label{eq:thm3-01}
    \Jp\leq (2d+3)\cdot C\cdot\Jsp, 
\end{equation}
where $C=\max\{1,2C_{\sigma}^2,2C_{\sigma}^4,5C_1^2,14C_2^2\}$, $C_1=\max\{B_{\sigma'}\cdot\max\{1,B_{\sigma'}\},C_{\sigma'}\cdot\max\{1,B_{\sigma'},C_{\sigma}\}\}$, and $C_2=\max\{B_{\sigma''}\cdot\max\{1,B_{\sigma'},C_{\sigma'},B_{\sigma'}^2,B_{\sigma'}C_{\sigma'}\},\max\{1,C_{\sigma'}\}\cdot\max\{C_{\sigma'},B_{\sigma'}C_{\sigma''}\}\}$.
\end{thm}

\begin{thm}\label{thm3-2}
For all $\{W_l,b_l\}_{l=1}^3$, there exists $\{a_1,a_2\}$, $\{d_{li}\}_{i=0,\ldots,d}^{l=1,2}$, and $\{q_{i}\}_{i=1,\ldots,d}$, such that
\begin{equation}
\Jsp\leq\Jp.
\end{equation}
\end{thm}

The above theorems show the consistency between the LySep model and the PINN model for parabolic equations. The proof is similar to that of Theorem \ref{thm1}-\ref{thm2}, so we do not present it here. 

We further design an alternating direction algorithm to implement the LySep model. The minimization process of $\Jsp$ is similar to the Algorithm \ref{alg:01} for the elliptic equation. Therefore, we omit the details here and directly present the algorithm (Algorithm \ref{alg:02}).

\begin{algorithm}
\DontPrintSemicolon
\SetAlCapFnt{\fontsize{11pt}{18pt}}
\fontsize{10pt}{12pt}\selectfont
\KwIn{data $X$, $Y$; number of iterations $N_k$; learning rate $\tau$}
\KwOut{a feasible solution $\{W_l,b_l,a_l,d_{li},q_{i}\}$.}
\Begin{initialize $\{W_l,b_l\}_{l=1}^3$, $\{a_l,\{d_{li}\}_{i=0}^d\}_{l=1}^2$, and $\{q_{i}\}_{i=1}^d$\;
\For{$k = 0,\cdots,N_k-1$}{
\For{$i = 1,\cdots,d+1$}{
$W_1(:,i)\leftarrow W_1(:,i)-\tau\nabla_{W_1(:,i)}\Jsp$
}
Solve the linear system for $b_1$ obtained in \eqref{eq:Js-parabolic}\;
$a_1\leftarrow a_1-\tau\nabla_{a_1}\Jsp$\;
\For{$i = 0,\cdots,d$}{
$d_{1i}\leftarrow d_{1i}-\tau\nabla_{d_{1i}}\Jsp$
}
$W_2\leftarrow W_2-\tau\nabla_{W_2}\Jsp$\;
Solve the linear system for $b_2$ obtained in \eqref{eq:Js-parabolic}\;
$a_2\leftarrow a_2-\tau\nabla_{a_2}\Jsp$\;
\For{$i = 0,\cdots,d$}{
$d_{2i}\leftarrow d_{2i}-\tau\nabla_{d_{2i}}\Jsp$
}
\For{$i = 1,\cdots,d$}{
$q_{i}\leftarrow q_{i}-\tau\nabla_{q_{i}}\Jsp$
}
Solve the linear system for $W_3$ obtained in \eqref{eq:Js-parabolic}\;
Solve the linear system for $b_3$ obtained in \eqref{eq:Js-parabolic}\;
}
return $\{W_l,b_l\}_{l=1}^3$, $\{a_l,\{d_{li}\}_{i=0}^d\}_{l=1}^2$, and $\{q_{i}\}_{i=1}^d$.
}
\caption{Solve $\min_{W_l,b_l,a_l,d_{li},q_{i}} \Jsp$}\label{alg:02}
\end{algorithm}

\subsubsection{Methods for hyperbolic equations}
Similar to parabolic PDEs, we use the chain rule and variable substitution for PINN $\Psi_{\h}(t,\bx;\theta)$ in the hyperbolic problem and add these new variables and their constraints to the loss function via weighted penalty terms. Then, we present the LySep loss function with auxiliary variables for problem \eqref{eq:Hyperbolic equation}:
\begin{multline}\label{eq:Js-hyberbolic}
\Jsh := \frac{1}{N}\Bigg[\Big\|K^{\h}*(W_3\sigma(a_2)+b_3\Bt)+K^{\h}_0*(W_3(\sigma'(a_2)*d_{2,0}))+\hat K^{\h}_0*(W_3(\sigma''(a_2)*d_{2,0}*d_{2,0}+\sigma'(a_2)*q_{2,0}))\\
-\sum_{i=1}^dK^{\h}_i*(W_3(\sigma'(a_2)*d_{2i}))-\sum_{i=1}^d\hat K^{\h}_i*(W_3(\sigma''(a_2)*d_{2i}*d_{2i}+\sigma'(a_2)*q_{i}))-Y\Big\|_\tF^2
+\omega_{a_1}^1\|W_1X_T+b_1\Bone^\top-a_1\|_{\mathcal{D}_{a_1}^1}^2\\
+\sum_{k=2}^3\sum_{i=0}^d\omega_{a_1}^{ki}\|W_1X_T+b_1\Bone^\top-a_1\|_{\mathcal{D}_{a_1}^{ki}}^2
+\omega_{a_2}^1\|W_2\sigma(a_1)+b_2\Bone^\top-a_2\|_{\mathcal{D}_{a_2}^1}^2
+\sum_{i=0}^d\omega_{a_2}^{2i}\|W_2\sigma(a_1)+b_2\Bone^\top-a_2\|_{\mathcal{D}_{a_2}^{2i}}^2\\
+\sum_{k=1}^2\sum_{i=0}^d\Big(\omega_{d_{1i}}^{k}\|W_1(:,i+1)\Bt-d_{1i}\|_{\mathcal{D}_{d_{1i}}^{k}}^2+\omega_{d_{2i}}^{k}\|W_2(\sigma'(a_1)*d_{1i})-d_{2i}\|_{\mathcal{D}_{d_{2i}}^{k}}^2\Big)
+\sum_{i=0}^d\omega_{q_i}\|W_2(\sigma''(a_1)*d_{1i}*d_{1i})-q_{i}\|_\tF^2\Bigg],
\end{multline}
where 
\begin{align*}
&K^{\h}=2\Big((\hat X-1)-d\hat{T}_0*c(X)-\hat{T}_0*\sum_{i=1}^dX(i,:)*\pxi c(X)\Big);
\quad K^{\h}_0=4T_0*(\hat X-1);\\
&K^{\h}_i=\hat K^{\h}_0*\pxi c(X)+4\hat{T}_0*X(i,:)*c(X);\quad 
\hat K^{\h}_0=\hat{T}_0*(\hat X-1);\quad 
\hat K^{\h}_i=c(X)*\hat K^{\h}_0,
\end{align*}
the specific expressions for the weight terms are as follows
\begin{align*}
&\omega_{a_1}^1=\omega_{d_{1i}}^{2}=\|W_3\|_\tF^2\|W_2\|_\tF^2;\quad \omega_{a_1}^{2i}=\omega_{a_2}^{2i}=\omega_{a_1}^1\|W_1(:,i+1)\|_\tF^2;\quad
\omega_{a_1}^{3i}=\omega_{a_1}^{2i}\|W_2\|_\tF^2;\\
&\omega_{a_2}^1=\omega_{d_{2i}}^{1}=\|W_3\|_\tF^2;\quad
\omega_{d_{1i}}^{1}=|\hat K_i^{\h}|_\infty^2(\omega_{a_1}^{2i}+\omega_{a_1}^{3i});\quad
\omega_{d_{2i}}^{2}=|\hat K_i^{\h}|_\infty^2\omega_{a_1}^{2i};\quad
\omega_{q_i}=|\hat K_i^{\h}|_\infty^2\omega_{a_2}^{1},
\end{align*}
and the specific expressions for the diagonal matrices are as follows
\begin{align*}
&\mathcal{D}_{a_1}^1(n,n)=\Big(\mathcal{D}_{a_2}^1(n,n)^2+\sum_{i=0}^d\Big(|K_i^{\h}|_\infty^2\|d_{1i}[n]\|_\tF^2+\mathcal{D}_{a_1}^{2i}(n,n)^2\|d_{2i}[n]\|_\tF^2\Big)\Big)^{\frac 12};\\
&\mathcal{D}_{a_1}^{2i}(n,n)=|\hat K_i^{\h}|_\infty\|d_{1i}[n]\|_\tF;\quad
\mathcal{D}_{a_1}^{3i}(n,n)=\Big(\mathcal{D}_{a_1}^{2i}(n,n)^2+\mathcal{D}_{a_2}^{2i}(n,n)^2\Big)^{\frac 12};\\
&\mathcal{D}_{a_2}^1(n,n)=\Big(|K^{\h}|_{\infty}^2+\sum_{i=0}^d(|K_i^{\h}|_\infty^2\|d_{2i}[n]\|_\tF^2+|\hat K^{\h}_i|_{\infty}^2\|q_{i}[n]\|_\tF^2)\Big)^{\frac 12};\\ 
&\mathcal{D}_{a_2}^{2i}(n,n)=|\hat K_i^{\h}|_\infty\|d_{2i}[n]\|_\tF;\quad
\mathcal{D}_{d_{1i}}^{1}=\mathcal{D}_{d_{2i}}^{2}=I;\\
&\mathcal{D}_{d_{1i}}^{2}(n,n)=\Big(|K^{\h}_i|_{\infty}^2+\mathcal{D}_{a_1}^{3i}(n,n)^2\Big)^{\frac 12};\quad
\mathcal{D}_{d_{2i}}^{1}(n,n)=\Big(|K^{\h}_i|_{\infty}^2+\mathcal{D}_{a_2}^{2i}(n,n)^2\Big)^{\frac 12},
\end{align*}
for $i=0,\ldots,d$ and $n=1,\ldots,N$.

Then, the LySep model for hyperbolic equations is given by
\begin{equation}\label{model_LySep3}
    \min_{W_l,b_l,a_l,d_{li},q_{i}}\Jsh.
\end{equation}

The consistency between $\Jsh$ and $\Jh$ is guaranteed by the following theorems, whose proof is similar to that of Theorem \ref{thm1}-\ref{thm2} and thus not presented here. 
\begin{thm}\label{thm4}
   Suppose $\sigma$, $\sigma'$ and $\sigma''$ are Lipschitz continuous and bounded functions, i.e., $|\sigma(z_1)-\sigma(z_2)|\leq C_{\sigma}|z_1-z_2|$, $|\sigma'(z_1)-\sigma'(z_2)|\leq C_{\sigma'}|z_1-z_2|$, $|\sigma''(z_1)-\sigma''(z_2)|\leq C_{\sigma''}|z_1-z_2|$, $|\sigma(z_1)|\leq B_{\sigma}$, $|\sigma'(z_1)|\leq B_{\sigma'}$ and $|\sigma''(z_1)|\leq B_{\sigma''}$, for some $C_{\sigma}, C_{\sigma'}, C_{\sigma''}, B_{\sigma'}, B_{\sigma''}>0$, any $z_1,z_2\in\mathbb{R}$. Given $\Jh$ and $\Jsh$ defined in \eqref{eq:J-hyberbolic} and \eqref{eq:Js-hyberbolic}, respectively. Then, for all $\{W_l,b_l\}_{l=1}^3$, $\{a_1,a_2\}$, $\{d_{li}\}_{i=0,\ldots,d}^{l=1,2}$, and $\{q_{i}\}_{i=0,\ldots,d}$, it holds that
\begin{equation}
    \Jh\leq 2(d+2)\cdot C\cdot\Jsh, 
\end{equation}
where $C=\max\{1,2C_{\sigma}^2,2C_{\sigma}^4,5C_1^2,14C_2^2\}$, $C_1=\max\{B_{\sigma'}\cdot\max\{1,B_{\sigma'}\},C_{\sigma'}\cdot\max\{1,B_{\sigma'},C_{\sigma}\}\}$, and $C_2=\max\{B_{\sigma''}\cdot\max\{1,B_{\sigma'},C_{\sigma'},B_{\sigma'}^2,B_{\sigma'}C_{\sigma'}\},\max\{1,C_{\sigma'}\}\cdot\max\{C_{\sigma'},B_{\sigma'}C_{\sigma''}\}\}$. 
\end{thm}
\begin{thm}\label{thm4-1}
For all $\{W_l,b_l\}_{l=1}^3$, there exists $\{a_1,a_2\}$, $\{d_{li}\}_{i=0,\ldots,d}^{l=1,2}$, and $\{q_{i}\}_{i=0,\ldots,d}$, such that
\begin{equation}
\Jsh\leq\Jh.
\end{equation}
\end{thm}

We directly give the algorithm of the LySep model for the hyperbolic equation (Algorithm \ref{alg:03}).

\begin{algorithm}
\DontPrintSemicolon
\SetAlCapFnt{\fontsize{11pt}{18pt}}
\fontsize{10pt}{12pt}\selectfont
\KwIn{data $X$, $Y$; number of iterations $N_k$; learning rate $\tau$}
\KwOut{a feasible solution $\{W_l,b_l,a_l,d_{li},q_{i}\}$.}
\Begin{initialize $\{W_l,b_l\}_{l=1}^3$, $\{a_l,\{d_{li}\}_{i=0}^d\}_{l=1}^2$, and $\{q_{i}\}_{i=0}^d$\;
\For{$k = 0,\cdots,N_k-1$}{
\For{$i = 1,\cdots,d+1$}{
$W_1(:,i)\leftarrow W_1(:,i)-\tau\nabla_{W_1(:,i)}\Jsh$
}
Solve the linear system for $b_1$ obtained in \eqref{eq:Js-hyberbolic}\;
$a_1\leftarrow a_1-\tau\nabla_{a_1}\Jsh$\;
\For{$i = 0,\cdots,d$}{
$d_{1i}\leftarrow d_{1i}-\tau\nabla_{d_{1i}}\Jsh$
}
$W_2\leftarrow W_2-\tau\nabla_{W_2}\Jsh$\;
Solve the linear system for $b_2$ obtained in \eqref{eq:Js-hyberbolic}\;
$a_2\leftarrow a_2-\tau\nabla_{a_2}\Jsh$\;
\For{$i = 0,\cdots,d$}{
$d_{2i}\leftarrow d_{2i}-\tau\nabla_{d_{2i}}\Jsh$
}
\For{$i = 0,\cdots,d$}{
$q_{i}\leftarrow q_{i}-\tau\nabla_{q_{i}}\Jsh$
}
Solve the linear system for $W_3$ obtained in \eqref{eq:Js-hyberbolic}\;
Solve the linear system for $b_3$ obtained in \eqref{eq:Js-hyberbolic}\;
}
return $\{W_l,b_l\}_{l=1}^3$, $\{a_l,\{d_{li}\}_{i=0}^d\}_{l=1}^2$, and $\{q_{i}\}_{i=0}^d$.
}
\caption{Solve $\min_{W_l,b_l,a_l,d_{li},q_{i}} \Jsh$}\label{alg:03}
\end{algorithm}

\section{Numerical Experiments}\label{sec_experiments}
In this section, we implement the PINN and LySep models to solve elliptic, parabolic, and hyperbolic equations. Details about the experimental settings are listed below. 
\begin{itemize}
  \item {\em Environment for PINN.}
  
  PINNs are implemented in Python with the PyTorch library. The optimization is solved by the vanilla gradient descent method with fine-tuned decaying learning rates. 
  \item {\em Environment for LySep.}
  
  LySep models are solved by Algorithms \ref{alg:01}-\ref{alg:03}, respectively. They are implemented in Matlab, where the subroutines {\tt mldivide} (i.e., \textbackslash ) or {\tt mrdivide} (i.e., /) are called to solve the linear least squares systems.
  
  \item {\em Initialization of variables.}
  
  The network parameters $\{W_l, b_l\}$ are randomly initialized with uniform distribution by
  \begin{equation}
  W_l, b_l\sim U(-M^{-1/2},M^{-1/2}).
  \end{equation}
  
  \item {\em Datasets and error evaluation.}
  
  The training feature vectors $\{x_n\}$ are chosen as Halton quasirandom points in the problem domain $\Omega$. Then the training dataset is formed by $\{(x_n,y_n)\}$ with $y_n=f(x_n)$. And we generate an extra set of Halton points $\{x'_n\}$ in $\Omega$, and let $\{(x'_n,y'_n)\}$ be the testing set, where $y'_n=f(x'_n)$. Note $\{x'_n\}$ is not overlapped with $\{x_n\}$. We compute the solution error, i.e. the $\ell^2$ error between the approximate network $\psi$ and the true PDE solution $u$,
  \begin{equation}
  \hEltwo=\left(\frac{\sum_{n=1}^{N}|\psi(x'_n;\theta)-u(x'_n)|^2}{\sum_{n=1}^{N}|u(x'_n)|^2}\right)^\frac{1}{2},
  \end{equation}
  to evaluate the generalization performance.
  
  \item {\em Randomness.}
  
  To alleviate the effect of randomness from initialization, we repeat every individual test using 10 different random seeds. This is equivalent to taking different initial guesses for the optimization. We present the mean and standard deviation (shown as ``mean $\pm$ standard deviation'') of the results of our numerical experiments in the table below. From the data in the table, it can be seen that the standard deviations are always dominated by the means, so our realization is numerically stable and convincing.

\end{itemize}

\subsection{Elliptic Problem}
In the first part of the numerical experiments, we employ the PINN and LySep models to solve the problem \eqref{eq:Elliptic equation}. We use Algorithm \ref{alg:01} to solve the LySep model. The activation function is set as the $\sin$ activation, which is Lipschitz continuous up to second-order derivatives, thus satisfying the hypothesis of Theorem \ref{thm1}.
\subsubsection{Two-dimensional case}
In this first example, we consider the elliptic problem \eqref{eq:Elliptic equation} in the 2-D domain $\Omega=\{\bx \in\mathbb{R}^2,\|\bx\|_2\leq 1\}$ and set the true solution
\begin{equation*}
     u(\bx) = (\exp(x_1^2+x_2^2-1)-1)(\sin(x_1)+\sin(x_2)).
\end{equation*}
The coefficient function is set as $c(\bx)=x_1^2+x_2^2$. The sizes of training and testing sets are $1000$ and $350$, respectively. For LySep model, we minimize the loss function $\Jse$ over $2\times10^3$ iterations, yielding a feasible solution $\{W_l,b_l,a_l,d_{li},q_{i}\}$. Additionally, we also compute the mean squared loss $\Je$ of the learned parameters $\{W_l,b_l\}$, verifying the consistency between $\Je$ and $\Jse$. For comparison, we use the gradient descent algorithm for the PINN model with $2\times10^3$ iterations to obtain a reference solution. To test the performance of various network sizes, we implement the algorithms with four depth-width combinations $(L,M)=(3, 30)$, $(3,50)$, $(3, 80)$, and $(3, 100)$. 

\begin{figure}[!ht]
\centering
\subfloat[$L=3,M=30$]{
\includegraphics[scale=0.5]{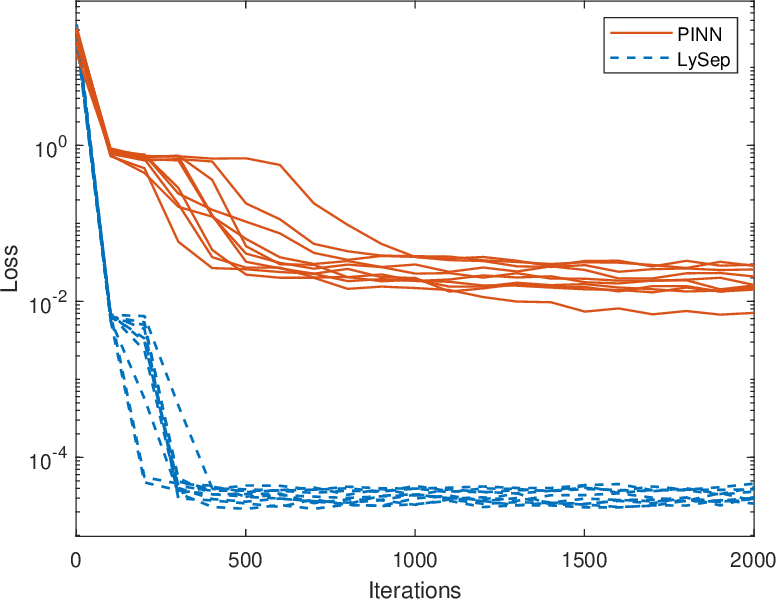}}
\subfloat[$L=3,M=50$]{
\includegraphics[scale=0.5]{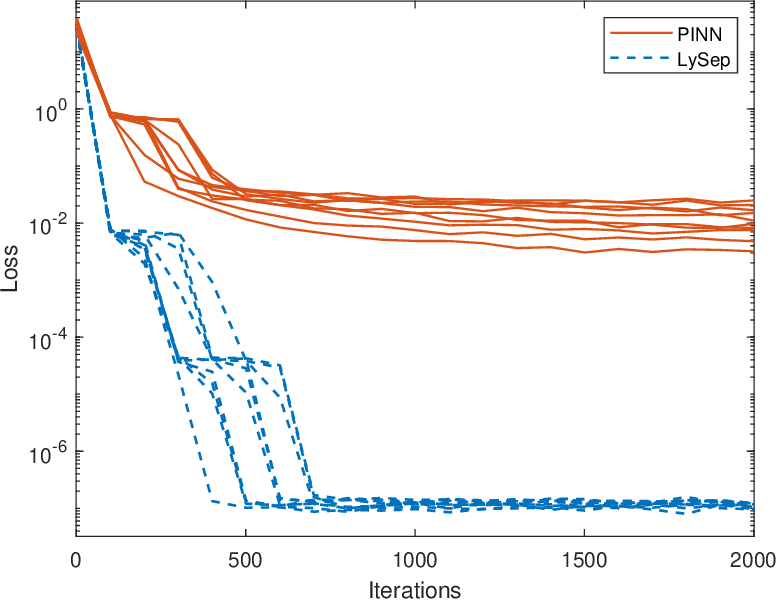}}\\
\subfloat[$L=3,M=80$]{
\includegraphics[scale=0.5]{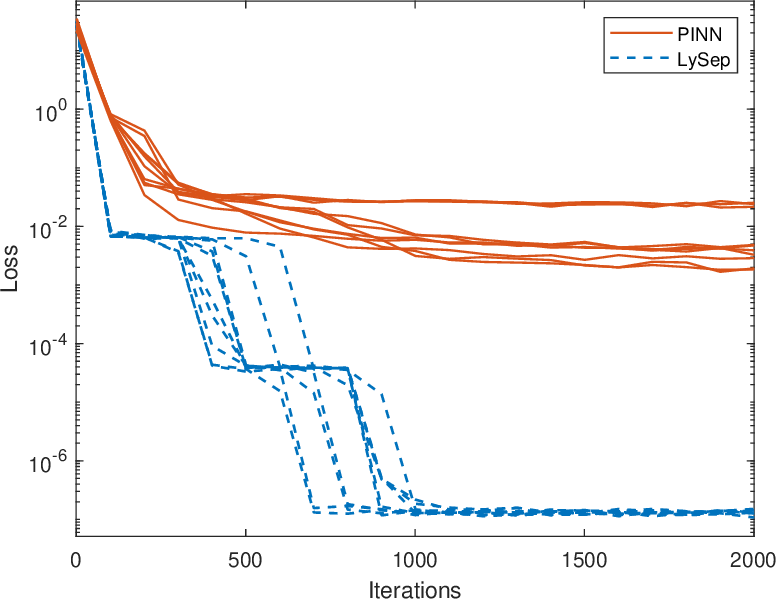}}
\subfloat[$L=3,M=100$]{
\includegraphics[scale=0.5]{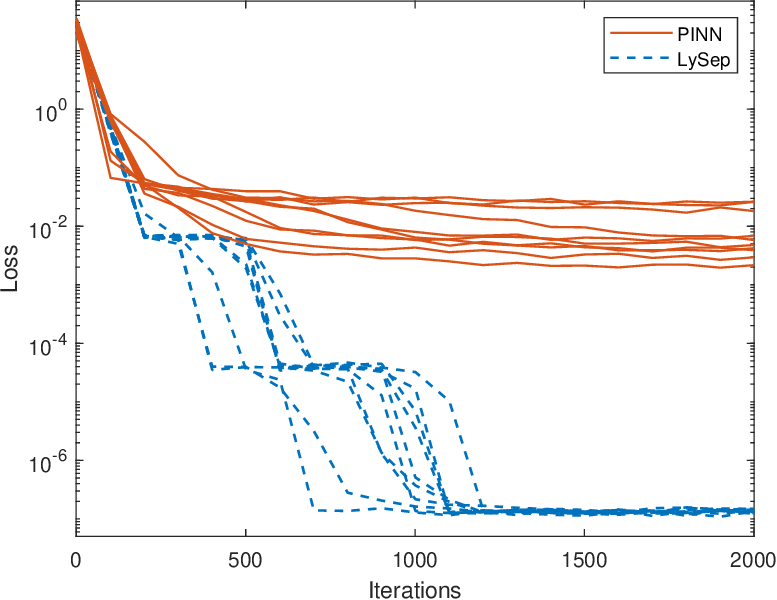}}\\
\caption{\em The loss $\Je$ versus iterations in solving the 2-D elliptic example with the PINN and LySep models (10 tests).}
\label{Fig_example1_loss}
\end{figure}

First, we investigate the decreasing trend of the loss function during the iteration process. Fig.\ref{Fig_example1_loss} plots $\Je$ versus the number of iterations for PINN and LySep models over all random seeds, respectively. It can be observed that the loss curves corresponding to the PINN model initially decrease, then stagnate for a short period, decrease again, and finally remain at $O(10^{-2})$ as the number of iterations increases. In addition, as the width of the network increases, the time of stagnation becomes shorter. This implies that in the process of optimizing the PINN model using the gradient descent algorithm, it may stagnate in bad regions. In contrast, for the LySep model, the loss $\Je$ keeps escaping from bad regions throughout the training process. At network size of $(L,M)=(3,30)$, the finial loss stagnates in $O(10^{-4})$. At other network sizes, the final losses are stagnant in the $O(10^{-8})$ region. Moreover, we can observe that the the curves of loss are more clustered with lower variance when optimizing the LySep model, which implies that the LySep model is more robust to random initialization.

\begin{figure}[!ht]
\centering
\subfloat[$L=3,M=30$]{
\includegraphics[scale=0.5]{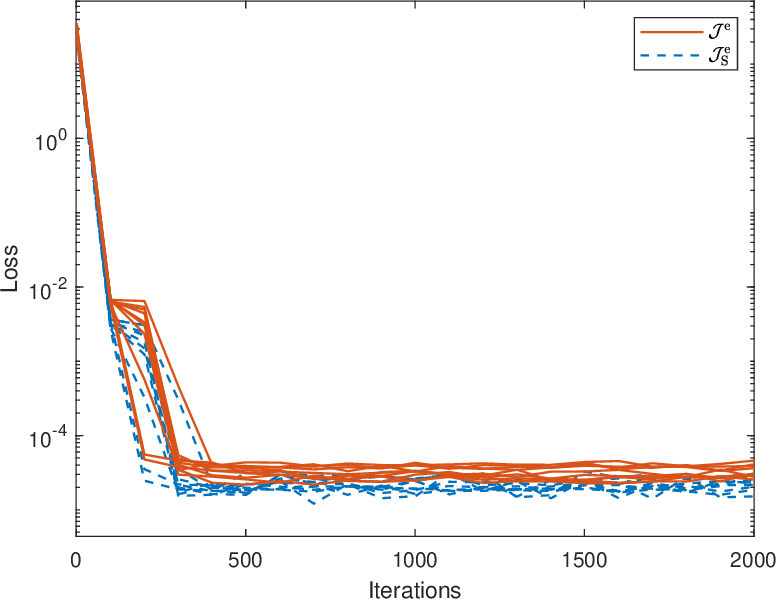}}
\subfloat[$L=3,M=50$]{
\includegraphics[scale=0.5]{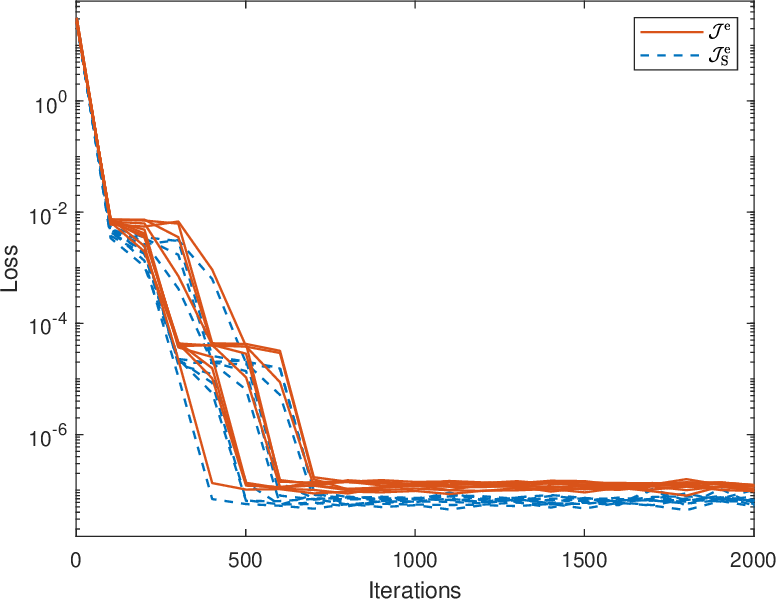}}\\
\subfloat[$L=3,M=80$]{
\includegraphics[scale=0.5]{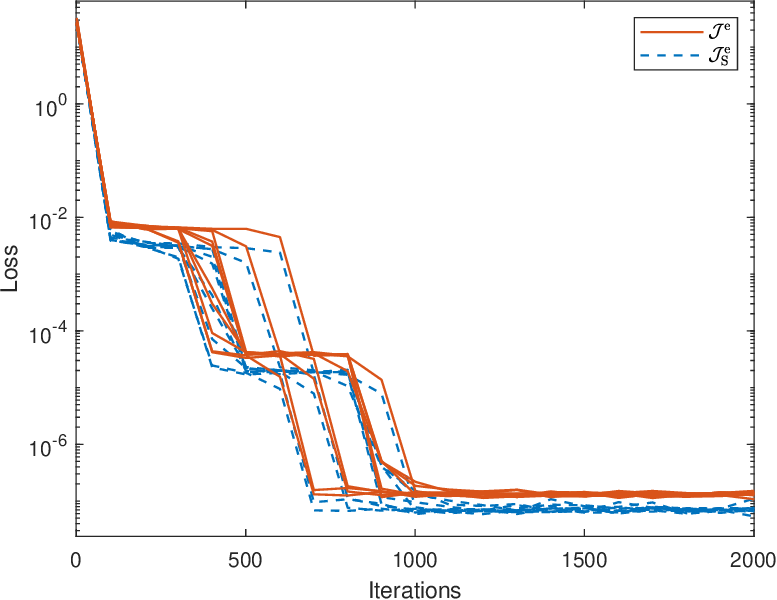}}
\subfloat[$L=3,M=100$]{
\includegraphics[scale=0.5]{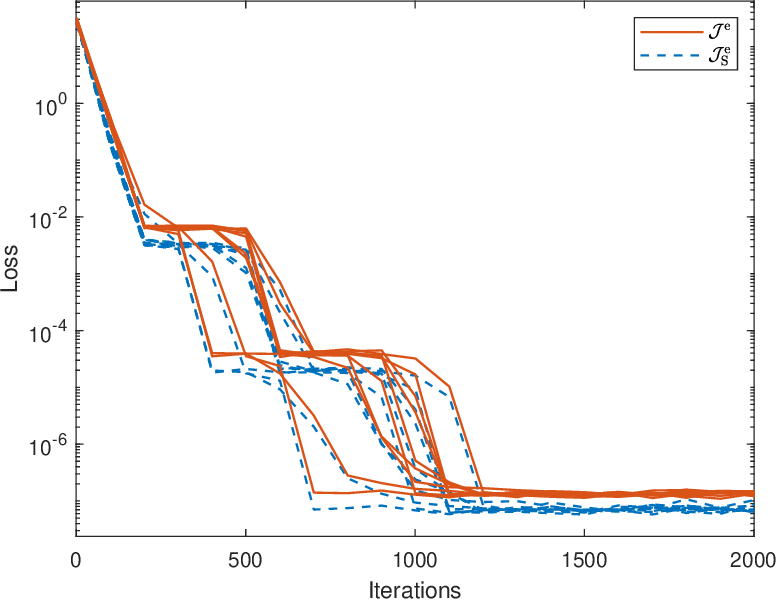}}\\
\caption{\em The losses $\Jse$ and $\Je$ versus iterations in solving the 2-D elliptic example with the LySep model (10 tests).}
\label{Fig_example1_loss_ORloss}
\end{figure}

Then, in Fig.\ref{Fig_example1_loss_ORloss}, we show the curves of loss functions $\Jse$ and $\Je$ of training the LySep model with the number of iterations under all seeds. It can be seen that the curves of $\Jse$ and $\Je$ decrease in parallel, which is consistent with Theorems \ref{thm1} and \ref{thm2}. More precisely, Theorem \ref{thm1} shows that the upper limit of $\Je$ is $2(d+1)\cdot C \cdot \Jse$. In this example, $C = 14$, so $\Je$ should not exceed $28(d+1)\cdot\Jse$ in theory. Our numerical results show that the inequality $\Je\leq 2(d+1)\cdot C\cdot\Jse$ always holds when training the LySep model under all seeds.

Finally,  we give the mean and standard deviation of the loss and error for both models after 2000 iterations in Table \ref{Tab:example1_bestresult}. 
From the table, it can be seen that for different network sizes, the loss $\Je$ and error $\hEltwo$ obtained from PINN are always higher than LySep. As the network size increases, the system becomes larger and more difficult to solve; the loss $\Je$ and error $\hEltwo$ obtained from PINN both increase slightly, but $\Jse$, $\Je$, and $\hEltwo$ obtained from LySep are still decreasing. At a network size of $(L,M)=(3,30)$, the test error of LySep is $O(10^{-3})$. At other network sizes, LySep's test error stays at $O(10^{-4})$.

\begin{table}[!ht]
\fontsize{9}{10.5}\selectfont
\setlength{\tabcolsep}{6pt} 
\centering
\begin{tabular}{clccc}
  \toprule
$(L,M)$ & Models & Actual loss & $\Je$ &  $\hEltwo$ \\\hline
\multirow{2}{*}{$(3,30)$} 
& PINN & \multicolumn{2}{c}{ 1.88e-02$\pm$6.78e-03}   & 5.68e-02$\pm$1.01e-02 \\
& LySep &  2.36e-05$\pm$4.72e-06 & 3.35e-05$\pm$6.32e-06  & 2.20e-03$\pm$2.81e-04 \\\hline
\multirow{2}{*}{$(3,50)$} 
& PINN & \multicolumn{2}{c}{1.22e-02$\pm$6.68e-03}   &  4.25e-02$\pm$1.57e-02\\
& LySep &  6.61e-08$\pm$1.38e-08 & 1.12e-07$\pm$1.03e-08  & 1.01e-04$\pm$1.15e-05 \\\hline
\multirow{2}{*}{$(3,80)$} 
& PINN & \multicolumn{2}{c}{9.42e-03$\pm$9.38e-03}  & 3.08e-02$\pm$2.42e-02  \\
& LySep & 7.39e-08$\pm$1.39e-08 & 1.34e-07$\pm$1.12e-08  & 1.10e-04$\pm$1.12e-05 \\\hline
\multirow{2}{*}{$(3,100)$} 
& PINN & \multicolumn{2}{c}{1.01e-02$\pm$9.09e-03}  & 3.19e-02$\pm$2.32e-02 \\
& LySep & 7.65e-08$\pm$1.22e-08 & 1.35e-07$\pm$9.35e-08  &1.14e-04$\pm$1.15e-05 \\
\bottomrule
\end{tabular}
\captionsetup{width=\textwidth}
\caption{\em The final actual losses, associated mean squared losses $\Je$ and testing errors $\hEltwo$ of the 2-D elliptic example.}
\label{Tab:example1_bestresult}
\end{table}

\subsubsection{High-dimensional case}
In this case, we consider the elliptic problem \eqref{eq:Elliptic equation} with the coefficient function $c(\bx)=\frac{1}{d}\sum_{i=1}^dx_i^2$ in 10-D unit ball $\Omega=\{\bx \in\mathbb{R}^{10},\|\bx\|_2\leq 1\}$. The true solution is set to be $u(\bx) = \sin\Big(\frac 1d\Big(\sum_{i=1}^dx_i^2-1\Big)\Big)$. To accommodate the high-dimensional case, we enlarge the training dataset to $2\times 10^3$. The network size is taken as $(L,M)=(3,20)$, $(3,50)$, $(3,100)$, and $(3,150)$. 

Phenomena similar to the preceding case are observed. We present the mean and standard deviation of the loss $\Je$ and training error $\hEltwo$ obtained by LySep and PINN models for all seeds, as well as the loss $\Jse$ obtained by LySep model in Table \ref{Tab:example2}. From the table, both $\Je$ and $\hEltwo$ obtained from the PINN model are higher than the LySep model at different network sizes. 
In addition, as the network size increases, both $\Je$ and $\hEltwo$ obtained from the PINN model slightly increase, but those obtained from the LySep model decrease. At a network size of $(L,M)=(3,20)$, the test error obtained by LySep model is $O(10^{-3})$. At other network sizes, the test error remains at $O(10^{-4})$. Moreover, the $\Jse$ and $\Je$ obtained from the LySep model decrease simultaneously as the network width increases. Theorem \ref{thm1} shows that the upper limit of $\Je$ is $2(d+1)\cdot C\cdot\Jse$ . In this example, we also get $C = 14$, so $\Je$ should also not exceed $28(d+1)\cdot\Jse$ in theory. Our numerical results show that the $\Je\leq 2(d+1)\cdot C\cdot\Jse$ relation in the LySep model always holds under all seed conditions in high-dimensional case.

\begin{table}[!ht]
\fontsize{9}{10.5}\selectfont
\setlength{\tabcolsep}{6pt} 
\centering
\begin{tabular}{clccc}
  \toprule
$(L,M)$ & Models & Actual loss & $\Je$ &  $\hEltwo$ \\\hline
\multirow{2}{*}{$(3,20)$} 
& PINN & \multicolumn{2}{c}{ 2.82e-06$\pm$1.93e-06}   &  7.29e-03$\pm$2.50e-03 \\
& LySep & 1.09e-07$\pm$7.98e-08 & 1.25e-07$\pm$ 9.67e-08 & 1.12e-03$\pm$5.51e-04 \\\hline
\multirow{2}{*}{$(3,50)$} 
& PINN & \multicolumn{2}{c}{ 2.67e-06$\pm$8.32e-07}   &  8.99e-03$\pm$1.42e-03 \\
& LySep & 2.72e-09$\pm$2.36e-09 & 2.96e-09$\pm$2.89e-09 & 3.66e-04$\pm$9.02e-05 \\\hline
\multirow{2}{*}{$(3,100)$} 
& PINN & \multicolumn{2}{c}{ 4.98e-06$\pm$1.69e-06}   &  1.38e-02$\pm$2.34e-03 \\
& LySep & 9.85e-10$\pm$2.23e-10 & 1.22e-09$\pm$ 3.96e-10 & 2.42e-04$\pm$2.94e-05 \\\hline
\multirow{2}{*}{$(3,150)$} 
& PINN & \multicolumn{2}{c}{ 1.05e-05$\pm$2.19e-06}   &  2.00e-02$\pm$2.16e-03 \\
& LySep & 5.38e-10$\pm$5.34e-11 & 7.41e-10$\pm$ 9.21e-11 & 1.84e-04$\pm$4.15e-05  \\
\bottomrule
\end{tabular}
\captionsetup{width=\textwidth}
\caption{\em The final actual losses, associated mean squared losses $\Je$ and testing errors $\hEltwo$ of the 10-D elliptic example.}
\label{Tab:example2}
\end{table}

\subsection{Parabolic Problem}
In the third example, we use the PINN and LySep models to solve the parabolic problem \eqref{eq:Parabolic equation} with $c(\bx)=\frac{1}{d}\sum_{i=1}^dx_i+2$ and let the true solution be
\begin{equation*}
     u(t,\bx) = \Big(\exp\Big(-\frac{t}{d}\Big)-1\Big)\sin\Big(\frac{1}{d}\Big(\sum_{i=1}^dx_i^2-1\Big)\Big)\sum_{i=1}^d\cos\Big(\frac{x_i}{\sqrt{d}}\Big),
\end{equation*}
where $d=5$. We set $\Omega=\{\bx \in\mathbb{R}^{5},\|\bx\|_2\leq 1\}$ and $[0,T]=[0,1]$. The activation function is also set as the $\sin$ activation. Here, we choose $N=2\times 10^3$ points to form the training dataset. We implement the algorithms with four depth-width combinations $(L,M)=(3, 20)$, $(3,50)$, $(3, 80)$, and $(3, 100)$. Similarly, we present the value of $\Jsp$, $\Jp$, and $\hEltwo$ from the LySep model in Table \ref{Tab:example3} under four different network sizes. For comparison, we also present the actual loss $\Jp$ and test error $\hEltwo$ from the PINN model.

From Table \ref{Tab:example3}, it can be seen that $\Jp$ obtained by the PINN model is lower than the LySep model at $(L,M)=(3,20)$. However, as the network size increases, $\Jp$ obtained by PINN model increases from $O(10^{-5})$ to $O(10^{-4})$, and $\hEltwo$ increases from $O(10^{-3})$ to $O(10^{-2})$. Comparatively, $\Jp$ obtained by LySep model decreases from $O(10^{-4})$ to $O(10^{-5})$ and $\hEltwo$ decreases from $O(10^{-3})$ to $O(10^{-4})$. In addition, $\Jsp$ and $\Jp$ obtained by the LySep model are consistent. Theorems \ref{thm3}-\ref{thm3-2} show that the upper bound of $\Jp$ is $(2d+3)\cdot C\cdot\Jsp$. In this example, $C = 14$, so theoretically $\Jp$ should also not exceed $14(2d+3)\cdot\Jsp$, which always holds in our numerical results.
\begin{table}[!ht]
\fontsize{9}{10.5}\selectfont
\setlength{\tabcolsep}{6pt} 
\centering
\begin{tabular}{clccc}
  \toprule
$(L,M)$ & Models & Actual loss & $\Jp$ &  $\hEltwo$ \\\hline
\multirow{2}{*}{$(3,20)$} 
& PINN & \multicolumn{2}{c}{ 5.86e-05$\pm$1.64e-05}   &  3.49e-03$\pm$9.76e-04 \\
& LySep &  1.55e-04$\pm$4.56e-05 &   1.55e-04$\pm$4.53e-05 &  3.14e-03$\pm$7.66e-03 \\\hline
\multirow{2}{*}{$(3,50)$} 
& PINN & \multicolumn{2}{c}{ 2.17e-04$\pm$7.46e-05}   &  5.68e-03$\pm$1.47e-03 \\
& LySep &  2.95e-05$\pm$3.95e-06 &  2.97e-05$\pm$3.99e-06  &  1.06e-03$\pm$6.81e-05 \\\hline
\multirow{2}{*}{$(3,80)$} 
& PINN & \multicolumn{2}{c}{ 6.45e-04$\pm$3.41e-04}   &  1.04e-02$\pm$2.04e-03 \\
& LySep &  1.07e-05$\pm$1.02e-06 &  1.42e-05$\pm$1.17e-06  &  8.68e-04$\pm$1.42e-04 \\\hline
\multirow{2}{*}{$(3,100)$} 
& PINN & \multicolumn{2}{c}{ 8.88e-04$\pm$5.47e-04}   &  1.19e-02$\pm$3.81e-03 \\
& LySep &  1.03e-05$\pm$6.87e-07 &  1.40e-05$\pm$9.40e-07  &  8.44e-04$\pm$1.58e-04 \\
\bottomrule
\end{tabular}
\captionsetup{width=\textwidth}
\caption{\em The final actual losses, associated mean squared losses $\Jp$ and testing errors $\hEltwo$ of the 5-D parabolic example.}
\label{Tab:example3}
\end{table}

\subsection{Hyperbolic Problem}
In the fourth example, we solve \eqref{eq:Hyperbolic equation} in the 5-D domain $\Omega=\{\bx \in\mathbb{R}^{5},\|\bx\|_2\leq 1\}$ and $[0,T]=[0,1]$. The coefficient function set as $c(\bx)=\frac{1}{d}\sum_{i=1}^dx_i+2$. We let the true solution be
\begin{equation*}
     u(t,\bx) = \Big(\exp\Big(-\frac{t^2}{d}\Big)-1\Big)\sin\Big(\frac{1}{d}\Big(\sum_{i=1}^dx_i^2-1\Big)\Big)\sum_{i=1}^d\cos\Big(\frac{x_i}{\sqrt{d}}\Big).
\end{equation*}
The activation function is also set as the $\sin$ activation. The number of iterations, the training set, and the depth-width combinations of neural networks are the same as in the third example. From Table \ref{Tab:example4}, it can be obtained that the numerical results from the LySep model are more accurate than the PINN model. Also, the consistency between $\Jh$ and $\Jsh$ claimed in Theorems \ref{thm4}-\ref{thm4-1} is validated numerically.
\begin{table}[!ht]
\fontsize{9}{10.5}\selectfont
\setlength{\tabcolsep}{6pt} 
\centering
\begin{tabular}{clccc}
  \toprule
$(L,M)$ & Models & Actual loss & $\Jh$ &  $\hEltwo$ \\\hline
\multirow{2}{*}{$(3,20)$} 
& PINN & \multicolumn{2}{c}{ 4.85e-05$\pm$1.29e-05}   &  7.04e-03$\pm$2.68e-03 \\
& LySep & 5.78e-05$\pm$1.11e-05 & 9.31e-05$\pm$2.14e-05  & 6.69e-03$\pm$1.29e-03 \\\hline
\multirow{2}{*}{$(3,50)$} 
& PINN & \multicolumn{2}{c}{ 2.09e-04$\pm$1.19e-04}   &  8.08e-03$\pm$2.48e-03 \\
& LySep & 1.78e-05$\pm$4.70e-06 & 2.74e-05$\pm$7.20e-06  & 2.81e-03$\pm$4.11e-04 \\\hline
\multirow{2}{*}{$(3,80)$} 
& PINN & \multicolumn{2}{c}{ 3.27e-04$\pm$2.25e-04}   &  1.13e-02$\pm$3.22e-03 \\
& LySep & 5.80e-06$\pm$1.03e-06 & 1.01e-05$\pm$1.81e-06  & 9.68e-04$\pm$1.15e-04 \\\hline
\multirow{2}{*}{$(3,100)$} 
& PINN & \multicolumn{2}{c}{ 2.75e-03$\pm$6.39e-03}   &  1.87e-02$\pm$1.73e-02 \\
& LySep & 5.24e-06$\pm$4.19e-07 & 9.78e-06$\pm$1.02e-06  & 9.40e-04$\pm$1.42e-04 \\
\bottomrule
\end{tabular}
\captionsetup{width=\textwidth}
\caption{\em The final actual losses, associated mean squared losses $\Jh$ and testing errors $\hEltwo$ of the 5-D hyperbolic example.} 
\label{Tab:example4}
\end{table}

\section{Conclusion}\label{sec:concludsion}
This work investigates deep learning models for second-order PDEs. Due to the high non-convexity of the mean square loss in deep learning, it is usually difficult to find good minima with common optimizers. We propose a novel LySep model, which introduces auxiliary variables to reformulate the physics-informed neural networks. Specifically, the LySep model decouples the outputs of all layers of PINNs as well as their derivatives, decomposing the deep network into a sequence of shallow networks. We present the specific forms of the LySep models for elliptic, parabolic, and hyperbolic PDEs, and provide theoretical analyses demonstrating the consistency between the LySep model and the original PINN model. In high-dimensional numerical examples, we compare the losses/errors computed by the LySep and PINN models, which indicates that the LySep model can achieve smaller final loss and higher solution accuracy, particularly for wider networks. Furthermore, the numerical results validate our theory of consistency; namely, optimizing the LySep loss indeed decreases the original mean square loss.

For future research, the LySep framework could be extended to other neural networks, such as convolutional neural networks (CNNs). The LySep strategy may provide new perspectives on optimizing CNN models. Specifically, one can explore how to separate the convolutional layer and pooling layer by making use of their sparse structure and shared parameters, thereby decomposing the deep network into shallow sub-networks. Additionally, one can polish the alternating direction algorithms by reordering the updates of variables in a better way, exploring more efficient optimizers for non-convex variables (e.g., Adam), or proposing hierarchical update strategies for the variables that are nearly convex. 

%\appendix
\begin{appendix}
\section{The proof of Theorem 2.1 and 2.2}\label{proof}
\subsection{The proof of Theorem 2.1}
Bringing \eqref{eq:Phi} into $\Je$ and using the triangle inequality yields
\begin{align}\label{eq:thm_e_02}
    &\Big\|\sum_{i=1}^d\pxi(c(X)*\pxi\psi_{\e}(X))-Y\Big\|_\tF\nonumber\\
    \leq& \Big\|K^{\e}*(W_3\sigma(a_2)+b_3\Bt)+\sum_{i=1}^dK^{\e}_i*(W_3(\sigma'(a_2)*d_{2i}))+\hat K^{\e}*\sum_{i=1}^dW_3(\sigma''(a_2)*d_{2i}*d_{2i}+\sigma'(a_2)*q_{i})-Y\Big\|_\tF\nonumber\\
&+|K^{\e}|_{\infty}C_{\sigma}\|W_3\|_\tF\|\R_2\|_\tF+\sum_{i=1}^d|K^{\e}_i|_{\infty}\|W_3\|_\tF\|\T_{2i}\|_\tF+|\hat K^{\e}|_{\infty}\sum_{i=1}^d\|W_3\|_\tF\|\Q_{2i}\|_\tF,
\end{align}    
where $|\cdot|_\infty$ is the infinite norm of the vector, $\R_2=W_2\sigma(W_1X+b_1\Bt)+b_2\Bt-a_2$, 
$\T_{2i}=\sigma'(W_2\sigma(W_1X+b_1\Bt)+b_2\Bt)*\pxi(W_2\sigma(W_1X+b_1\Bt)+b_2\Bt)-\sigma'(a_2)*d_{2i}$, 
and $\Q_{2i}=\sigma''(W_2\sigma(W_1X+b_1\Bt)+b_2\Bt)*\pxi(W_2\sigma(W_1X+b_1\Bt)+b_2\Bt)*\pxi(W_2\sigma(W_1X+b_1\Bt)+b_2\Bt)+\sigma'(W_2\sigma(W_1X+b_1\Bt)+b_2\Bt)*\pxii(W_2\sigma(W_1X+b_1\Bt)+b_2\Bt)-(\sigma''(a_2)*d_{2i}*d_{2i}+\sigma'(a_2)*q_{i})$.

The detailed procedure for handling $\|\R_2\|_\tF$ and $\|\T_{2i}\|_\tF$ has been presented in \cite{Liu2025}. For the sake of completeness and readability of the proof, we briefly describe the main steps here. For the term $\|\R_2\|_\tF$, we apply the triangle inequality and the Lipschitz continuity of the activation function $\sigma$ to derive
\begin{align}\label{eq:thm_e_R2}
    \|\R_2\|_\tF
    \leq \|W_2\|_\tF\|\sigma(W_1X+b_1\Bt)-\sigma(a_1)\|_\tF+\|W_2\sigma(a_1)+b_2\Bt-a_2\|_\tF\no\\
    \leq C_{\sigma}\|W_2\|_\tF\|W_1X+b_1\Bt-a_1\|_\tF+\|W_2\sigma(a_1)+b_2\Bt-a_2\|_\tF.
\end{align}
For the term $\T_{2i}$, $\|\T_{2i}\|_\tF^2=\sum_{n=1}^N\|\T_{2i}[n]\|_\tF^2$ holds, where $\T_{2i}[n]$ means the $n$-th column of the matrix $\T_{2i}$. We also use the triangle inequality, chain rule, and the Lipschitz continuity of $\sigma'$, and obtain
\begin{align}\label{eq:thm_e_T2_1}
    \|\T_{2i}[n]\|_\tF\leq&|\sigma'(W_2\sigma(W_1x_n+b_1)+b_2)|_\infty\|\mathcal{S}_{2i}[n]\|_\tF+\|d_{2i}[n]\|_\tF\|\sigma'(W_2\sigma(W_1x_n+b_1)+b_2)-\sigma'(a_2[n])\|_\tF\no\\
    \leq& B_{\sigma'}\|\mathcal{S}_{2i}[n]\|_\tF+C_{\sigma'}\|d_{2i}[n]\|_\tF\|R_2[n]\|_\tF,
\end{align}
where $\mathcal{S}_{2i}[n] = \pxi(W_2\sigma(W_1x_n+b_1)+b_2)-d_{2i}[n]$. Similarly, we can derive
\begin{align}\label{eq:thm_e_S2}
    \|\mathcal{S}_{2i}[n]\|_\tF\leq\|W_2\|_\tF\|\T_{1i}[n]\|_\tF+\|W_2(\sigma'(a_1[n])*d_{1i}[n])-d_{2i}[n]\|_\tF,
\end{align}
and
\begin{align}\label{eq:thm_e_T1}
    \|\T_{1i}[n]\|_\tF=&\|\sigma'(W_1x_n+b_1)*W_1(:,i)-\sigma'(a_1[n])*d_{1i}[n]\|_\tF\no\\
    \leq|&\sigma'(W_1x_n+b_1)|_\infty\|W_1(:,i)-d_{1i}[n]\|_\tF+\|d_{1i}[n]\|_\tF\|\sigma'(W_1x_n+b_1)-\sigma'(a_1[n])\|_\tF\no\\
    \leq& B_{\sigma'}\|W_1(:,i)-d_{1i}[n]\|_\tF+C_{\sigma'}\|d_{1i}[n]\|_\tF\|W_1x_n+b_1-a_1[n]\|_\tF.
\end{align}
Combining \eqref{eq:thm_e_S2}-\eqref{eq:thm_e_T1} and bringing the result with \eqref{eq:thm_e_R2} into \eqref{eq:thm_e_T2_1} yields
\begin{multline}\label{eq:thm_e_T2_2}
    \|\T_{2i}[n]\|_\tF\leq C_1\Big(\|W_2\|_\tF\|W_1(:,i)-d_{1i}[n]\|_\tF+\|W_2(\sigma'(a_1[n])*d_{1i}[n])-d_{2i}[n]\|_\tF\\
    +\|W_2\|_\tF(\|d_{1i}[n]\|_\tF+\|d_{2i}[n]\|_\tF)\|W_1x_n+b_1-a_1[n]\|_\tF  +\|d_{2i}[n]\|_\tF\|W_2\sigma(a_1[n])+b_2-a_2[n]\|_\tF\Big),
\end{multline}
and
\begin{multline}\label{eq:thm_e_T2_3}
    \|\T_{2i}\|_\tF^2\leq5C_1^2\sum_{n=1}^N\Big[\|W_2\|_\tF^2\|W_1(:,i)-d_{1i}[n]\|_\tF^2
    +\|W_2(\sigma'(a_1[n])*d_{1i}[n])-d_{2i}[n]\|_\tF^2\\
    +\|W_2\|_\tF^2(\|d_{1i}[n]\|_\tF^2+\|d_{2i}[n]\|_\tF^2)\|W_1x_n+b_1-a_1[n]\|_\tF^2
    +\|d_{2i}[n]\|_\tF^2\|W_2\sigma(a_1[n])+b_2-a_2[n]\|_\tF^2\Big],
\end{multline}
where $C_1=\max\{B_{\sigma'}\cdot\max\{1,B_{\sigma'}\},C_{\sigma'}\cdot\max\{1,B_{\sigma'},C_{\sigma}\}\}$.

Then, for the term $\Q_{2i}$, $\|\Q_{2i}\|_\tF^2=\sum_{n=1}^N\|\Q_{2i}[n]\|_\tF^2$ also holds. Using the triangle inequality, chain rule, and the Lipschitz continuity of $\sigma''$, we obtain
\begin{multline}\label{eq:thm_e_Q2_1}
    \|\Q_{2i}[n]\|_\tF\leq \|\sigma''(W_2\sigma(W_1x_n+b_1)+b_2)*\pxi(W_2\sigma(W_1x_n+b_1)+b_2)
    *\pxi(W_2\sigma(W_1x_n+b_1)+b_2)\\-\sigma''(a_2[n])*d_{2i}[n]*d_{2i}[n]\|_\tF
    +\|\sigma'(W_2\sigma(W_1x_n+b_1)+b_2)*\pxii(W_2\sigma(W_1x_n+b_1)+b_2)-\sigma'(a_2[n])*q_{i}[n]\|_\tF\\
    \leq|\pxi(W_2\sigma(W_1x_n+b_1)+b_2)|_\infty\|\sigma''(W_2\sigma(W_1x_n+b_1)+b_2)*\pxi(W_2\sigma(W_1x_n+b_1)+b_2)
    -\sigma''(a_2[n])*d_{2i}[n]\|_\tF\\+|\sigma''(a_2[n])|_\infty\|d_{2i}[n]\|_\tF\|\mathcal{S}_{2i}[n]\|_\tF
    +|\sigma'(W_2\sigma(W_1x_n+b_1)+b_2)|_\infty\|\pxii(W_2\sigma(W_1x_n+b_1)+b_2)-q_{i}[n]\|_\tF\\
    +\|q_{i}[n]\|_\tF\|\sigma'(W_2\sigma(W_1x_n+b_1)+b_2)-\sigma'(a_2[n])\|_\tF\\
    \leq B_{\sigma'}\|W_2\|_\tF\|W_1(:,i)\|_\tF\Big(|\sigma''(W_2\sigma(W_1x_n+b_1)+b_2)|_\infty\|\mathcal{S}_{2i}[n]\|_\tF\\
    +\|d_{2i}[n]\|_\tF\|\sigma''(W_2\sigma(W_1x_n+b_1)+b_2)-\sigma''(a_2[n])\|_\tF\Big)
    +B_{\sigma''}\|d_{2i}[n]\|_\tF\|\mathcal{S}_{2i}[n]\|_\tF
    +B_{\sigma'}\|W_2\|_\tF\|\Q_{1i}[n]\|_\tF\\
    +B_{\sigma'}\|W_2(\sigma''(a_1[n])*d_{1i}[n]*d_{1i}[n])-q_{i}[n]\|_\tF+C_{\sigma'}\|q_{i}[n]\|_\tF\|R_2[n]\|_\tF\\
    \leq B_{\sigma''}(B_{\sigma'}\|W_2\|_\tF\|W_1(:,i)\|_\tF+\|d_{2i}[n]\|_\tF)\|\mathcal{S}_{2i}[n]\|_\tF\\
    +(B_{\sigma'}C_{\sigma''}\|W_2\|_\tF\|W_1(:,i)\|_\tF\|d_{2i}[n]\|_\tF+C_{\sigma'}\|q_{i}[n]\|_\tF)\|\R_2[n]\|_\tF\\
    +B_{\sigma'}\|W_2\|_\tF\|\Q_{1i}[n]\|_\tF
    +B_{\sigma'}\|W_2(\sigma''(a_1[n])*d_{1i}[n]*d_{1i}[n])-q_{i}[n]\|_\tF,
\end{multline}
where $\Q_{1i}[n]=\sigma''(W_1x_n+b_1)*W_1(:,i)*W_1(:,i)-\sigma''(a_1[n])*d_{1i}[n]*d_{1i}[n]$. Similar to the derivation of $\|\Q_{2i}[n]\|_\tF$, we can get the following results about $\|\Q_{1i}[n]\|_\tF$:
\begin{multline}\label{eq:thm_e_Q1}
    \|\Q_{1i}[n]\|_\tF\leq\|W_1(:,i)\|_\tF\|\sigma''(W_1x_n+b_1)*W_1(:,i)-\sigma''(a_1[n])*d_{1i}[n]\|_\tF\\
     +|\sigma''(a_1[n])|_\infty\|d_{1i}[n]\|_\tF\|W_1(:,i)-d_{1i}[n]\|_\tF\\
     \leq\|W_1(:,i)\|_\tF|\sigma''(W_1x_n+b_1)|_\infty\|W_1(:,i)-d_{1i}[n]\|_\tF
     +\|W_1(:,i)\|_\tF\|d_{1i}[n]\|_\tF\|\sigma''(W_1x_n+b_1)-\sigma''(a_1[n])\|_\tF\\
     +B_{\sigma''}\|d_{1i}[n]\|_\tF\|W_1(:,i)-d_{1i}[n]\|_\tF\\
     \leq B_{\sigma''}(\|W_1(:,i)\|_\tF+\|d_{1i}[n]\|_\tF)\|W_1(:,i)-d_{1i}[n]\|_\tF
     +C_{\sigma''}\|W_1(:,i)\|_\tF\|d_{1i}[n]\|_\tF\|W_1x_n+b_1-a_1[n]\|_\tF.
\end{multline}
Substituting \eqref{eq:thm_e_R2}, \eqref{eq:thm_e_S2} and \eqref{eq:thm_e_Q1} into \eqref{eq:thm_e_Q2_1}, we can obtain
\begin{multline}\label{eq:thm_e_Q2_2}
    \|\Q_{2i}[n]\|_\tF\leq C_2\Big[\|W_2\|_\tF(\|W_1(:,i)\|_\tF(\|W_2\|_\tF+1)+\|d_{1i}[n]\|_\tF+\|d_{2i}[n]\|_\tF)\|W_1(:,i)-d_{1i}[n]\|_\tF\\
    +\|W_2\|_\tF\Big(\|W_2\|_\tF\|W_1(:,i)\|_\tF(\|d_{1i}[n]\|_\tF+\|d_{2i}[n]\|_\tF)
    +\|d_{1i}[n]\|_\tF\|d_{2i}[n]\|_\tF\\
    +\|W_1(:,i)\|_\tF\|d_{1i}[n]\|_\tF+\|q_{i}[n]\|_\tF\Big)\|W_1x_n+b_1-a_1[n]\|_\tF\\
    +(\|W_2\|_\tF\|W_1(:,i)\|_\tF+\|d_{2i}[n]\|_\tF)\|W_2(\sigma'(a_1[n])*d_{1i}[n])-d_{2i}[n]\|_\tF\\
    +(\|W_2\|_\tF\|W_1(:,i)\|_\tF\|d_{2i}[n]\|_\tF+\|q_{i}[n]\|_\tF)\|W_2\sigma(a_1[n])+b_2-a_2[n]\|_\tF\\
    +\|W_2(\sigma''(a_1[n])*d_{1i}[n]*d_{1i}[n])-q_{i}[n]\|_\tF\Big],
\end{multline}
and
\begin{multline}\label{eq:thm_e_Q2_3}
    \|\Q_{2i}\|_\tF^2\leq 14 C_2^2\sum_{n=1}^N\Big[\|W_2\|_\tF^2(\|W_1(:,i)\|_\tF^2(\|W_2\|_\tF^2+1)+\|d_{1i}[n]\|_\tF^2
    +\|d_{2i}[n]\|_\tF^2)\|W_1(:,i)-d_{1i}[n]\|_\tF^2\\
    +\|W_2\|_\tF^2\Big(\|W_2\|_\tF^2\|W_1(:,i)\|_\tF^2(\|d_{1i}[n]\|_\tF^2+\|d_{2i}[n]\|_\tF^2)+\|d_{1i}[n]\|_\tF^2\|d_{2i}[n]\|_\tF^2\\
    +\|W_1(:,i)\|_\tF^2\|d_{1i}[n]\|_\tF^2+\|q_{i}[n]\|_\tF^2\Big)\|W_1x_n+b_1-a_1[n]\|_\tF^2\\
    +(\|W_2\|_\tF^2\|W_1(:,i)\|_\tF^2+\|d_{2i}[n]\|_\tF^2)\|W_2(\sigma'(a_1[n])*d_{1i}[n])-d_{2i}[n]\|_\tF^2\\
    +(\|W_2\|_\tF^2\|W_1(:,i)\|_\tF^2\|d_{2i}[n]\|_\tF^2+\|q_{i}[n]\|_\tF^2)\|W_2\sigma(a_1[n])+b_2-a_2[n]\|_\tF^2\\
    +\|W_2(\sigma''(a_1[n])*d_{1i}[n]*d_{1i}[n])-q_{i}[n]\|_\tF^2\Big],
\end{multline}
where $C_2=\max\{B_{\sigma''}\cdot\max\{1,B_{\sigma'},C_{\sigma'},B_{\sigma'}^2,B_{\sigma'}C_{\sigma'}\},\max\{1,C_{\sigma'}\}\cdot\max\{C_{\sigma'},B_{\sigma'}C_{\sigma''}\}\}$.
%$C_2=\max\{B_{\sigma'},B_{\sigma''}\cdot\max\{1,B_{\sigma'}\}\cdot\max\{1,B_{\sigma'},C_{\sigma'}\},\max\{1,C_{\sigma}\}\cdot\max\{C_{\sigma'},B_{\sigma'}C_{\sigma''}\}\}$.

Using the triangle inequality and bringing \eqref{eq:thm_e_R2}, \eqref{eq:thm_e_T2_3} and \eqref{eq:thm_e_Q2_3} into \eqref{eq:thm_e_02} leads to
\begin{multline}\label{eq:thm_e_03}
   \Big\|\sum_{i=1}^d\pxi(c(X)*\pxi\psi_{\e}(X))-Y\Big\|_\tF^2
\leq2(d+1)\cdot C\cdot\Bigg[\Big\|K^{\e}*(W_3\sigma(a_2)+b_3\Bt)\\
+\sum_{i=1}^dK^{\e}_i*(W_3(\sigma'(a_2)*d_{2i}))+\hat K^{\e}*\sum_{i=1}^dW_3(\sigma''(a_2)*d_{2i}*d_{2i}+\sigma'(a_2)*q_{i})-Y\Big\|_\tF^2\\
    +\sum_{n=1}^N\|W_3\|_\tF^2\|W_2\|_\tF^2\Big[|K^{\e}|_{\infty}^2+\sum_{i=1}^d\Big((|K^{\e}_i|_{\infty}^2+|\hat K^{\e}|_{\infty}^2\|W_2\|_\tF^2\|W_1(:,i)\|_\tF^2)\\
    (\|d_{1i}[n]\|_\tF^2+\|d_{2i}[n]\|_\tF^2)+|\hat K^{\e}|_{\infty}^2(\|d_{1i}[n]\|_\tF^2\|d_{2i}[n]\|_\tF^2+\|W_1(:,i)\|_\tF^2\|d_{1i}[n]\|_\tF^2\\
    +\|q_{i}[n]\|_\tF^2)\Big)\Big]\|W_1x_n+b_1-a_1[n]\|_\tF^2+\sum_{n=1}^N\|W_3\|_\tF^2\Big[|K^{\e}|_{\infty}^2+\sum_{i=1}^d\Big(|K^{\e}_i|_{\infty}^2\|d_{2i}[n]\|_\tF^2\\
    +|\hat K^{\e}|_{\infty}^2(\|W_2\|_\tF^2\|W_1(:,i)\|_\tF^2\|d_{2i}[n]\|_\tF^2+\|q_{i}[n]\|_\tF^2)\Big)\Big]\|W_2\sigma(a_1[n])+b_2-a_2[n]\|_\tF^2\\
    +\sum_{n=1}^N\sum_{i=1}^d\|W_3\|_\tF^2\|W_2\|_\tF^2\Big[|K^{\e}_i|_{\infty}^2+|\hat K^{\e}|_{\infty}^2\Big(\|W_1(:,i)\|_\tF^2(\|W_2\|_\tF^2+1)+\|d_{1i}[n]\|_\tF^2\\
    +\|d_{2i}[n]\|_\tF^2\Big)\Big]\|W_1(:,i)-d_{1i}[n]\|_\tF^2+\sum_{n=1}^N\sum_{i=1}^d\|W_3\|_\tF^2\Big[|K^{\e}_i|_{\infty}^2\\
    +|\hat K^{\e}|_{\infty}^2(\|W_2\|_\tF^2\|W_1(:,i)\|_\tF^2+\|d_{2i}[n]\|_\tF^2)\Big]\|W_2(\sigma'(a_1[n])*d_{1i}[n])-d_{2i}[n]\|_\tF^2\\
    +|\hat K^{\e}|_{\infty}^2\sum_{i=1}^d\|W_3\|_\tF^2\|W_2(\sigma''(a_1)*d_{1i}*d_{1i})-q_{i}\|_\tF^2\Bigg],
\end{multline}    
where $C=\max\{1,2C_{\sigma}^2,2C_{\sigma}^4,5C_1^2,14C_2^2\}$. \eqref{eq:thm_e_03} leads to the desired result given in \eqref{eq:thm1-01}.\qed

\subsection{The proof of Theorem 2.2}
It suffices to let $a_1 = W_1X+b_1\Bone^\top$, $a_2=W_2\sigma(a_1)+b_2\Bone^\top$, $d_{1i}=W_1(:,i)\Bone^\top$, $d_{2i}=W_2\left(\sigma'(a_1)*d_{1i}\right)$, and $q_{i}=W_2\left(\sigma''(a_1)*d_{1i}*d_{1i}\right)$ for $i=1,\ldots,d$. In this case the equality \eqref{eq:thm1-02} holds.\qed
\end{appendix}

\vspace{10pt}
\noindent{\fontsize{9.0pt}{\baselineskip}\selectfont \textbf{Funding} This work is supported by National Natural Science Foundation of China Major Research Plan (G0592370101).}

\vspace{10pt}
\noindent{\fontsize{9.0pt}{\baselineskip}\selectfont\textbf{Data availability} Data will be made available on request.}

\vspace{10pt}
\noindent{\fontsize{9.0pt}{\baselineskip}\selectfont\textbf{Declaration of competing interest} The authors declare that they have no known competing financial interests or personal relationships that could have appeared to influence the work reported in this paper.}

\bibliographystyle{model1-num-names}
\bibliography{refs}

\begin{thebibliography}{32}
\expandafter\ifx\csname natexlab\endcsname\relax\def\natexlab#1{#1}\fi
\providecommand{\url}[1]{\texttt{#1}}
\providecommand{\href}[2]{#2}
\providecommand{\path}[1]{#1}
\providecommand{\DOIprefix}{doi:}
\providecommand{\ArXivprefix}{arXiv:}
\providecommand{\URLprefix}{URL: }
\providecommand{\Pubmedprefix}{pmid:}
\providecommand{\doi}[1]{\href{http://dx.doi.org/#1}{\path{#1}}}
\providecommand{\Pubmed}[1]{\href{pmid:#1}{\path{#1}}}
\providecommand{\bibinfo}[2]{#2}
\ifx\xfnm\relax \def\xfnm[#1]{\unskip,\space#1}\fi
%Type = Article
\bibitem[{Yserentant(2005)}]{Yserentant2005}
\bibinfo{author}{H.~Yserentant},
\newblock \bibinfo{title}{Sparse grid spaces for the numerical solution of the electronic schr{\"o}dinger equation},
\newblock \bibinfo{journal}{Numer. Math.} \bibinfo{volume}{101} (\bibinfo{year}{2005}) \bibinfo{pages}{381--389}.
%Type = Article
\bibitem[{Lee et~al.(2002)Lee, Wang, and Newell}]{Lee2002}
\bibinfo{author}{T.~Lee}, \bibinfo{author}{F.~Wang}, \bibinfo{author}{R.~Newell},
\newblock \bibinfo{title}{Robust model-order reduction of complex biological processes},
\newblock \bibinfo{journal}{J. Process Control} \bibinfo{volume}{12} (\bibinfo{year}{2002}) \bibinfo{pages}{807--821}.
%Type = Article
\bibitem[{Ehrhardt and R.~E~Mickens(2008)}]{Ehrhardt2008}
\bibinfo{author}{M.~Ehrhardt}, \bibinfo{author}{R.~E. R.~E~Mickens},
\newblock \bibinfo{title}{A fast, stable and accurate numerical method for the {Black--Scholes} equation of {American} options},
\newblock \bibinfo{journal}{Int. J. Theor. Appl. Finance} \bibinfo{volume}{11} (\bibinfo{year}{2008}) \bibinfo{pages}{471--501}.
%Type = Article
\bibitem[{Cai et~al.(2021{\natexlab{a}})Cai, Mao, Wang, Yin, and Karniadakis}]{Cai2022}
\bibinfo{author}{S.~Cai}, \bibinfo{author}{Z.~Mao}, \bibinfo{author}{Z.~Wang}, \bibinfo{author}{M.~Yin}, \bibinfo{author}{G.~E. Karniadakis},
\newblock \bibinfo{title}{Physics-informed neural networks ({PINN}s) for fluid mechanics: {A} review},
\newblock \bibinfo{journal}{Acta Mech. Sin.}  (\bibinfo{year}{2021}{\natexlab{a}}) \bibinfo{pages}{1727--1738}.
%Type = Article
\bibitem[{Cai et~al.(2021{\natexlab{b}})Cai, Wang, Wang, Perdikaris, and Karniadakis}]{Cai2021}
\bibinfo{author}{S.~Cai}, \bibinfo{author}{Z.~Wang}, \bibinfo{author}{S.~Wang}, \bibinfo{author}{P.~Perdikaris}, \bibinfo{author}{G.~E. Karniadakis},
\newblock \bibinfo{title}{Physics-informed neural networks for heat transfer problems},
\newblock \bibinfo{journal}{J. Heat Transf.} \bibinfo{volume}{143} (\bibinfo{year}{2021}{\natexlab{b}}) \bibinfo{pages}{060801}.
%Type = Article
\bibitem[{Jagtap and Karniadakis(2020)}]{Jagtap2020}
\bibinfo{author}{A.~D. Jagtap}, \bibinfo{author}{G.~E. Karniadakis},
\newblock \bibinfo{title}{Extended physics-informed neural networks {(XPINNs)}: A generalized space-time domain decomposition based deep learning framework for nonlinear partial differential equations},
\newblock \bibinfo{journal}{Commun. Comput. Phys.} \bibinfo{volume}{28} (\bibinfo{year}{2020}).
%Type = Article
\bibitem[{Mao et~al.(2020)Mao, Jagtap, and Karniadakis}]{Mao2020}
\bibinfo{author}{Z.~Mao}, \bibinfo{author}{A.~D. Jagtap}, \bibinfo{author}{G.~E. Karniadakis},
\newblock \bibinfo{title}{Physics-informed neural networks for high-speed flows},
\newblock \bibinfo{journal}{Comput. Methods Appl. Mech. Engrg.} \bibinfo{volume}{360} (\bibinfo{year}{2020}) \bibinfo{pages}{112789}.
%Type = Article
\bibitem[{Pang et~al.(2019)Pang, Lu, and Karniadakis}]{Pang2019}
\bibinfo{author}{G.~Pang}, \bibinfo{author}{L.~Lu}, \bibinfo{author}{G.~E. Karniadakis},
\newblock \bibinfo{title}{{fPINNs}: Fractional physics-informed neural networks},
\newblock \bibinfo{journal}{SIAM J. Sci. Comput.} \bibinfo{volume}{41} (\bibinfo{year}{2019}) \bibinfo{pages}{A2603--A2626}.
%Type = Article
\bibitem[{Raissi et~al.(2019)Raissi, Perdikaris, and Karniadakis}]{Raissi2019}
\bibinfo{author}{M.~Raissi}, \bibinfo{author}{P.~Perdikaris}, \bibinfo{author}{G.~E. Karniadakis},
\newblock \bibinfo{title}{Physics-informed neural networks: {A} deep learning framework for solving forward and inverse problems involving nonlinear partial differential equations},
\newblock \bibinfo{journal}{J. Comput. Phys.} \bibinfo{volume}{378} (\bibinfo{year}{2019}) \bibinfo{pages}{686--707}.
%Type = Article
\bibitem[{Rao et~al.(2020)Rao, Sun, and Liu}]{Rao2020}
\bibinfo{author}{C.~Rao}, \bibinfo{author}{H.~Sun}, \bibinfo{author}{Y.~Liu},
\newblock \bibinfo{title}{Physics-informed deep learning for incompressible laminar flows},
\newblock \bibinfo{journal}{Theor. Appl. Mech. Lett.} \bibinfo{volume}{10} (\bibinfo{year}{2020}) \bibinfo{pages}{207--212}.
%Type = Article
\bibitem[{Yu(2018)}]{Yu2018}
\bibinfo{author}{B.~Yu},
\newblock \bibinfo{title}{The deep {Ritz} method: a deep learning-based numerical algorithm for solving variational problems},
\newblock \bibinfo{journal}{Commun. Math. Stat.} \bibinfo{volume}{6} (\bibinfo{year}{2018}) \bibinfo{pages}{1--12}.
%Type = Article
\bibitem[{Gu et~al.(2021)Gu, Yang, and Zhou}]{Gu2021-2}
\bibinfo{author}{Y.~Gu}, \bibinfo{author}{H.~Yang}, \bibinfo{author}{C.~Zhou},
\newblock \bibinfo{title}{Selectnet: {Self}-paced learning for high-dimensional partial differential equations},
\newblock \bibinfo{journal}{J. Comput. Phys.} \bibinfo{volume}{441} (\bibinfo{year}{2021}) \bibinfo{pages}{110444}.
%Type = Article
\bibitem[{Zang et~al.(2020)Zang, Bao, Ye, and Zhou}]{Zang2020}
\bibinfo{author}{Y.~Zang}, \bibinfo{author}{G.~Bao}, \bibinfo{author}{X.~Ye}, \bibinfo{author}{H.~Zhou},
\newblock \bibinfo{title}{Weak adversarial networks for high-dimensional partial differential equations},
\newblock \bibinfo{journal}{J. Comput. Phys.} \bibinfo{volume}{411} (\bibinfo{year}{2020}) \bibinfo{pages}{109409}.
%Type = Unpublished
\bibitem[{Bao et~al.(2024)Bao, Wang, and Zou}]{Bao2024}
\bibinfo{author}{G.~Bao}, \bibinfo{author}{D.~Wang}, \bibinfo{author}{B.~Zou}, \bibinfo{title}{{WANCO}: weak adversarial networks for constrained optimization problems}, \bibinfo{year}{2024}. \bibinfo{note}{Https://arxiv.org/abs/2407.03647}.
%Type = Inproceedings
\bibitem[{Gao et~al.(2023)Gao, Gu, and Ng}]{Gao2023}
\bibinfo{author}{Y.~Gao}, \bibinfo{author}{Y.~Gu}, \bibinfo{author}{M.~Ng},
\newblock \bibinfo{title}{Gradient descent finds the global optima of two-layer physics-informed neural networks},
\newblock in: \bibinfo{booktitle}{Proceedings of the 40th International Conference on Machine Learning}, \bibinfo{year}{2023}, pp. \bibinfo{pages}{10676--10707}.
%Type = Unpublished
\bibitem[{Luo and Yang(2020)}]{Luo2020}
\bibinfo{author}{T.~Luo}, \bibinfo{author}{H.~Yang}, \bibinfo{title}{Two-layer neural networks for partial differential equations: Optimization and generalization theory}, \bibinfo{year}{2020}. \bibinfo{note}{Https://arxiv.org/abs/2006.15733}.
%Type = Article
\bibitem[{Gu et~al.(2021)Gu, Wang, and Yang}]{Gu2021}
\bibinfo{author}{Y.~Gu}, \bibinfo{author}{C.~Wang}, \bibinfo{author}{H.~Yang},
\newblock \bibinfo{title}{Structure probing neural network deflation},
\newblock \bibinfo{journal}{J. Comput. Phys.} \bibinfo{volume}{434} (\bibinfo{year}{2021}) \bibinfo{pages}{110231}.
%Type = Article
\bibitem[{R{\"o}gnvaldsson(1994)}]{Rognvaldsson1994}
\bibinfo{author}{T.~R{\"o}gnvaldsson},
\newblock \bibinfo{title}{On langevin updating in multilayer perceptrons},
\newblock \bibinfo{journal}{Neural Comput.} \bibinfo{volume}{6} (\bibinfo{year}{1994}) \bibinfo{pages}{916--926}.
%Type = Inproceedings
\bibitem[{Erhan et~al.(2009)Erhan, Manzagol, Bengio, Bengio, and Vincent}]{Erhan2009}
\bibinfo{author}{D.~Erhan}, \bibinfo{author}{P.~A. Manzagol}, \bibinfo{author}{Y.~Bengio}, \bibinfo{author}{S.~Bengio}, \bibinfo{author}{P.~Vincent},
\newblock \bibinfo{title}{The difficulty of training deep architectures and the effect of unsupervised pre-training},
\newblock in: \bibinfo{booktitle}{Proceedings of the 12th International Conference on Artificial Intelligence and Statistics}, \bibinfo{year}{2009}, pp. \bibinfo{pages}{153--160}.
%Type = Unpublished
\bibitem[{Liu et~al.(2025)Liu, Gu, and Ng}]{Liu2025}
\bibinfo{author}{Y.~Liu}, \bibinfo{author}{Y.~Gu}, \bibinfo{author}{M.~K. Ng}, \bibinfo{title}{Deep learning optimization using self-adaptive weighted auxiliary variables}, \bibinfo{year}{2025}. \bibinfo{note}{Https://arxiv.org/abs/2504.21501}.
%Type = Article
\bibitem[{Zhang et~al.(2022)Zhang, Shen, and Yang}]{Zhang2022}
\bibinfo{author}{S.~Zhang}, \bibinfo{author}{Z.~Shen}, \bibinfo{author}{H.~Yang},
\newblock \bibinfo{title}{Deep {N}etwork {A}pproximation: {Achieving Arbitrary Accuracy with Fixed Number of Neurons}},
\newblock \bibinfo{journal}{J. Mach. Learn. Res.} \bibinfo{volume}{23} (\bibinfo{year}{2022}) \bibinfo{pages}{1--60}.
%Type = Article
\bibitem[{Jiao et~al.(2023)Jiao, Lai, Lu, Wang, Yang, and Yang}]{Jiao2023}
\bibinfo{author}{Y.~Jiao}, \bibinfo{author}{Y.~Lai}, \bibinfo{author}{X.~Lu}, \bibinfo{author}{F.~Wang}, \bibinfo{author}{J.~Z. Yang}, \bibinfo{author}{Y.~Yang},
\newblock \bibinfo{title}{Deep neural networks with relu-sine-exponential activations break curse of dimensionality in approximation on h{\"o}lder class},
\newblock \bibinfo{journal}{SIAM J. Math. Anal.} \bibinfo{volume}{55} (\bibinfo{year}{2023}) \bibinfo{pages}{3635--3649}.
%Type = Article
\bibitem[{Shen et~al.(2021{\natexlab{a}})Shen, Yang, and Zhang}]{Shen2021_1}
\bibinfo{author}{Z.~Shen}, \bibinfo{author}{H.~Yang}, \bibinfo{author}{S.~Zhang},
\newblock \bibinfo{title}{Neural network approximation: {Three} hidden layers are enough},
\newblock \bibinfo{journal}{Neural Netw.} \bibinfo{volume}{141} (\bibinfo{year}{2021}{\natexlab{a}}) \bibinfo{pages}{160--173}.
%Type = Article
\bibitem[{Shen et~al.(2021{\natexlab{b}})Shen, Yang, and Zhang}]{Shen2021_2}
\bibinfo{author}{Z.~Shen}, \bibinfo{author}{H.~Yang}, \bibinfo{author}{S.~Zhang},
\newblock \bibinfo{title}{Deep network with approximation error being reciprocal of width to power of square root of depth},
\newblock \bibinfo{journal}{Neural Comput.} \bibinfo{volume}{33} (\bibinfo{year}{2021}{\natexlab{b}}) \bibinfo{pages}{1005--1036}.
%Type = Inproceedings
\bibitem[{Barron(1992)}]{Barron1992}
\bibinfo{author}{A.~R. Barron},
\newblock \bibinfo{title}{Neural net approximation},
\newblock in: \bibinfo{booktitle}{Proceedings of the 7th Yale Workshop on Adaptive and Learning Systems}, \bibinfo{year}{1992}, pp. \bibinfo{pages}{69--72}.
%Type = Article
\bibitem[{Barron(1993)}]{Barron1993}
\bibinfo{author}{A.~R. Barron},
\newblock \bibinfo{title}{Universal approximation bounds for superpositions of a sigmoidal function},
\newblock \bibinfo{journal}{IEEE Trans. Inform. Theory} \bibinfo{volume}{39} (\bibinfo{year}{1993}) \bibinfo{pages}{930--945}.
%Type = Article
\bibitem[{Caragea et~al.(2023)Caragea, Petersen, and Voigtlaender}]{Caragea2023}
\bibinfo{author}{A.~Caragea}, \bibinfo{author}{P.~Petersen}, \bibinfo{author}{F.~Voigtlaender},
\newblock \bibinfo{title}{Neural network approximation and estimation of classifiers with classification boundary in a {Barron} class},
\newblock \bibinfo{journal}{Ann. Appl. Probab.} \bibinfo{volume}{33} (\bibinfo{year}{2023}) \bibinfo{pages}{3039--3079}.
%Type = Unpublished
\bibitem[{E et~al.(2020)E, Ma, Wojtowytsch, and Wu}]{E2020}
\bibinfo{author}{W.~E}, \bibinfo{author}{C.~Ma}, \bibinfo{author}{S.~Wojtowytsch}, \bibinfo{author}{L.~Wu}, \bibinfo{title}{Towards a mathematical understanding of neural network-based machine learning: what we know and what we don't}, \bibinfo{year}{2020}. \bibinfo{note}{Https://arxiv.org/abs/2009.10713}.
%Type = Article
\bibitem[{E et~al.(2022)E, Ma, and Wu}]{E2022}
\bibinfo{author}{W.~E}, \bibinfo{author}{C.~Ma}, \bibinfo{author}{L.~Wu},
\newblock \bibinfo{title}{The {Barron} space and the flow-induced function spaces for neural network models},
\newblock \bibinfo{journal}{Constr. Approx.} \bibinfo{volume}{55} (\bibinfo{year}{2022}) \bibinfo{pages}{369--406}.
%Type = Article
\bibitem[{Klusowski and Barron(2018)}]{Klusowski2018}
\bibinfo{author}{J.~M. Klusowski}, \bibinfo{author}{A.~R. Barron},
\newblock \bibinfo{title}{Approximation by combinations of {ReLU} and squared {ReLU} ridge functions with $\ell^1$ and $\ell^0$ controls},
\newblock \bibinfo{journal}{IEEE Trans. Inform. Theory} \bibinfo{volume}{64} (\bibinfo{year}{2018}) \bibinfo{pages}{7649--7656}.
%Type = Article
\bibitem[{Siegel and Xu(2020)}]{Siegel2020_1}
\bibinfo{author}{J.~W. Siegel}, \bibinfo{author}{J.~Xu},
\newblock \bibinfo{title}{Approximation rates for neural networks with general activation functions},
\newblock \bibinfo{journal}{Neural Netw.} \bibinfo{volume}{128} (\bibinfo{year}{2020}) \bibinfo{pages}{313--321}.
%Type = Article
\bibitem[{Siegel and Xu(2022)}]{Siegel2022}
\bibinfo{author}{J.~W. Siegel}, \bibinfo{author}{J.~Xu},
\newblock \bibinfo{title}{High-order approximation rates for shallow neural networks with cosine and {ReLU}$^k$ activation functions},
\newblock \bibinfo{journal}{Appl. Comput. Harmon. Anal.} \bibinfo{volume}{58} (\bibinfo{year}{2022}) \bibinfo{pages}{1--26}.

\end{thebibliography}
	
\end{document}